\documentclass[preprint,3p,times,twocolumn]{elsarticle}

\usepackage{amssymb}  
\usepackage{amsmath} 
\usepackage{moreverb,url}
\usepackage{dblfloatfix}
\usepackage[table]{xcolor}
\usepackage[colorlinks,bookmarksopen,bookmarksnumbered,citecolor=red,urlcolor=red]{hyperref}
\usepackage{graphics} 
\usepackage{graphicx}
\usepackage{epsfig} 
\usepackage{amsthm}

\usepackage{algorithm, algpseudocode}

\usepackage{xcolor}
\definecolor{Light0}{rgb}{0.98, 0.95, 0.99}
\definecolor{Light1}{rgb}{0.98, 0.95, 0.90}
\definecolor{Light2}{rgb}{0.98, 0.98, 0.93}
\definecolor{Light3}{rgb}{0.98, 0.98, 1}

\usepackage{float}
\usepackage{todonotes}

\usepackage{subfigure}
\usepackage{algpseudocode}
\usepackage{epstopdf}
\usepackage{multicol}
\usepackage[leftcaption]{sidecap}
\usepackage{booktabs}

\usepackage{arydshln}
\usepackage{enumitem}

\usepackage[nolist]{acronym} 

\newacro{vic}[VIC]{Variable Impedance Control}

\newacro{pi2}[PI\textsuperscript{2}]{Policy Improvement with Path Integrals}
\newacro{dmp}[DMP]{Dynamic Movement Primitive}
\newacro{seds}[SEDS]{Stable Estimator of Dynamical Systems}
\newacro{ilc}[ILC]{Iterative  learning  control}
\newacro{gmm}[GMM]{Gaussian Mixture Model}
\newacro{gmr}[GMR]{Gaussian Mixture Regression}
\newacro{ppc}[PPC]{Passivity-Preservation Control}
\newacro{tpgmm}[TP-GMM]{Task-Parameterized \ac{gmm}}
\newacro{lfd}[LfD]{Learning from Demonstration}
\newacro{lfhd}[LfD]{Learning-from-human-Demonstration}
\newacro{il}[IL]{Imitation Learning}
\newacro{wls}[WLS]{weighted least-squares}
\newacro{spd}[SPD]{Symmetric Positive Definite}
\newacro{emg}[EMG]{Electromyography}
\newacro{vil}[VIL]{Variable Impedance Learning}
\newacro{vilc}[VILC]{Variable Impedance Learning Control}
\newacro{ai}[AI]{Artificial Intelligent}
\newacro{gadmp}[$\mathcal{G}$-DMP]{Geometry-aware \ac{dmp}}
\newacro{promp}[ProMP]{Probabilistic Movement Primitives}
\newacro{kmp}[KMP]{Kernelized Movement Primitives}
\newacro{dof}[DoF]{Degree of Freedom}
\newacro{msd}[MSD]{virtual-Mass Spring-Damper}
\newacro{}[]{}
\newacro{}[]{}

\newcommand{\bm}[1]{\boldsymbol{\mathbf{#1}}}

\relax
\DeclareFontFamily{U}{eur}{\skewchar\font'177}
\DeclareFontShape{U}{eur}{m}{n}{%
	<-6> eurm5 <6-8> eurm7 <8-> eurm10}{}
\DeclareFontShape{U}{eur}{b}{n}{%
	<-6> eurb5 <6-8> eurb7 <8-> eurb10}{}
\DeclareSymbolFont{ugrf@m}{U}{eur}{m}{n}
\SetSymbolFont{ugrf@m}{bold}{U}{eur}{b}{n}
\DeclareSymbolFont{EUr}{U}{eur}{m}{n}
\DeclareSymbolFont{EUb}{U}{eur}{b}{n}

\newcommand{\dy}{\Dot{y}}

\newcommand{\yd}{{\bm{\dot{y}}}}
\newcommand{\ydd}{{\bm{\ddot{y}}}}
\newcommand{\Y}{{\bm{Y}}}

\newcommand{\bcalY}{\bm{\mathcal{Y}}}
\newcommand{\bcalYhat}{\bm{\hat{\mathcal{Y}}}}
\newcommand{\bcalYd}{\bm{\dot{\mathcal{Y}}}}
\newcommand{\bcalYdd}{\bm{\ddot{\mathcal{Y}}}}

\newcommand{\dz}{\Dot{z}}
\newcommand{\Z}{{\bm{Z}}}

\newcommand{\bcalZ}{\bm{\mathcal{Z}}}
\newcommand{\bcalZd}{\bm{\dot{\mathcal{Z}}}}

\newcommand{\p}{{\bm{p}}}
\newcommand{\PP}{{\bm{P}}}

\newcommand{\bcalP}{{\bm{\mathcal{P}}}}
\newcommand{\bcalPd}{\bm{\dot{\mathcal{P}}}}

\newcommand{\bcalPhat}{\bm{\hat{\mathcal{P}}}}

\newcommand{\bcalJ}{{\bm{\mathcal{J}}}}
\newcommand{\bcalJd}{\bm{\dot{\mathcal{J}}}}

\newcommand{\bcalV}{{\bm{\mathcal{V}}}}
\newcommand{\bcalVd}{\bm{\dot{\mathcal{V}}}}

\newcommand{\G}{{\bm{G}}}

\newcommand{\bcalGd}{\bm{\dot{\mathcal{G}}}}

\newcommand{\bcalQ}{{\bm{\mathcal{Q}}}}

\newcommand{\U}{{\bm{U}}}
\newcommand{\bcalU}{{\bm{\mathcal{U}}}}
\newcommand{\bcalH}{{\bm{\mathcal{H}}}}
\newcommand{\bcalHd}{\bm{\dot{\mathcal{H}}}}

\newcommand{\bcalK}{{\bm{\mathcal{K}}}}
\newcommand{\bcalD}{{\bm{\mathcal{D}}}}

\newcommand\manS[1]{\mathcal{S}^{#1}}
\newcommand\manR[1]{\mathcal{R}^{#1}}
\newcommand\manSO[1]{\mathcal{SO}\left(#1\right)}
\newcommand\manSE[1]{\mathcal{SE}\left(#1\right)}
\newcommand{\spd}[1]{{\mathcal{S}}_{++}^{#1}}
\newcommand{\sym}[1]{\mathcal{SYM}^{#1}}

\newcommand{\calM}{{\mathcal{M}}}
\newcommand{\calN}{{\mathcal{N}}}
\newcommand{\calL}{{\mathcal{L}}}

\newcommand{\calT}{{\mathcal{T}}}
\newcommand{\bcalF}{{\bm{\mathcal{F}}}}
\newcommand{\bcalW}{{\bm{\mathcal{W}}}}
\newcommand{\bcalR}{{\bm{\mathcal{R}}}}
\newcommand{\bcalC}{{\bm{\mathcal{C}}}}
\newcommand{\bcalCd}{\bm{\dot{\mathcal{C}}}}

\newcommand{\TPM}{{\mathcal{T}_\PP\mathcal{M}}}
\newcommand{\TPPM}{{\mathcal{T}_{\PP_1}\mathcal{M}}}

\newcommand{\TUUN}{{\mathcal{T}_{\U_1}\mathcal{N}}}

\newcommand{\boldfrak}[1]{\bm{\mathfrak{#1}}}

\newcommand{\aEXPmap}{\bm{A}=\text{Exp}_{\PP}(\bm{a})}
\newcommand{\aLOGmap}{\bm{a}= \text{Log}_{\PP}(\bm{A})}

\newcommand{\w}{\bm{w}}

\newcommand{\R}{\bm{R}}

\def\eQCoRL{$e\bcalQ^{dmp\  \text{\cite{koutras2020correct}}}$}
\def\eQdeQCoRL{$\norm{e\bcalQ^{demo}-e\bcalQ^{dmp\  \text{\cite{koutras2020correct}}}}$}
\def\eQGaDMP{$e\bcalQ^{\ac{gadmp}}$}
\def\eQdeQGaDMP{$\norm{e\bcalQ^{demo}-e\bcalQ^{\ac{gadmp}}}$}
\def\nuGaDMPrep{$\nu^{\ac{gadmp}_{reproduction}}$}
\def\nuGaDMPgol{$\nu^{\ac{gadmp}_{newGoal}}$}
\def\uxGaDMPrep{$\bm{u}_x^{\ac{gadmp}_{reproduction}}$}
\def\uxGaDMPgol{$\bm{u}_x^{\ac{gadmp}_{newGoal}}$}
\def\uyGaDMPrep{$\bm{u}_y^{\ac{gadmp}_{reproduction}}$}
\def\uyGaDMPgol{$\bm{u}_y^{\ac{gadmp}_{newGoal}}$}
\def\uzGaDMPrep{$\bm{u}_z^{\ac{gadmp}_{reproduction}}$}
\def\uzGaDMPgol{$\bm{u}_z^{\ac{gadmp}_{newGoal}}$}
\def\Upsdemo{$\bm{\Upsilon}^{demo}$}
\def\Upsdmp{$\bm{\Upsilon}^{\ac{dmp}}$}
\def\Pdmp{$\bcalP^{\ac{gadmp}}$}

\def\eKGaDMPrep{$e\bcalK^{\ac{gadmp}_{reproduction}}$}
\def\eKGaDMPgol{$e\bcalK^{\ac{gadmp}_{newGoal}}$}
\def\Koo{$\bcalK_{11}^{\ac{gadmp}_{newGoal}}$}
\def\Kot{$\bcalK_{12}^{\ac{gadmp}_{newGoal}}$}
\def\Ktt{$\bcalK_{22}^{\ac{gadmp}_{newGoal}}$}
\def\KGaDMPrep{$\bm{\bcalK}^{\ac{gadmp}_{reproduction}}$}

\newcommand\norm[1]{\left\lVert#1\right\rVert}
\newcommand\expm[1]{\text{expm}\left(#1\right)}

\newcommand\logm[1]{\text{logm}\left(#1\right)}
\newcommand{\Log}[2]{\text{Log}_{#1}\left(#2\right)}
\newcommand{\Exp}[2]{\text{Exp}_{#1}\left(#2\right)}

\newcommand{\TSM}[1]{{\mathcal{T}_{#1}\mathcal{M}}}

\newcommand{\trsp}{{^{\top}}}

\newcommand{\figref}[1]{Fig.~\hyperref[#1]{\ref*{#1}}}
\newcommand{\figsref}[1]{Figures~\hyperref[#1]{\ref*{#1}}}
\newcommand{\Figref}[1]{Figure~\hyperref[#1]{\ref*{#1}}}

\newcommand{\tabref}[1]{Tab.~\hyperref[#1]{\ref*{#1}}}
\newcommand{\Tabref}[1]{Table~\hyperref[#1]{\ref*{#1}}}
\newcommand{\secref}[1]{Sec.~\hyperref[#1]{\ref*{#1}}}
\newcommand{\Secref}[1]{Section~\hyperref[#1]{\ref*{#1}}}
\newcommand{\algoref}[1]{Alg.~\hyperref[#1]{\ref*{#1}}}
\newcommand{\Algoref}[1]{Algorithm~\hyperref[#1]{\ref*{#1}}}

\newcommand{\eg} {\textit{e.g.,}~} %
\newcommand{\ie} {\textit{i.e.,}~} %
\newcommand{\etal}{\MakeLowercase{{\textit{et~al}.\ }}}

\newcommand{\ordinal}[2]{#1$^{\text{#2}}$} %

\newlength{\Oldarrayrulewidth}
\newcommand{\Cline}[2]{%
  \noalign{\global\setlength{\Oldarrayrulewidth}{\arrayrulewidth}}%
  \noalign{\global\setlength{\arrayrulewidth}{#1}}\cline{#2}%
  \noalign{\global\setlength{\arrayrulewidth}{\Oldarrayrulewidth}}
}

\providecommand{\change }[1]{{\color{black}#1}}

\graphicspath{{figs/}}

\newtheorem{theorem}{\bf Theorem}
\newtheorem{remark}{\bf Remark}

\journal{Neurocomputing}

\begin{document}

\begin{frontmatter}
	
	
	\title{
		A Unified Formulation of Geometry-aware Discrete Dynamic Movement Primitives
	}


	\author[MU]{Fares~J.~Abu-Dakka\corref{mycorrespondingauthor}}
	\cortext[mycorrespondingauthor]{Corresponding author}
	\ead{fabudakka@mondragon.edu}
	
	\author[unitn]{Matteo~Saveriano}
	\ead{matteo.saveriano@unitn.it}
	
	\author[aalto]{Ville Kyrki}
	\ead{ville.kyrki@aalto.fi}
	
	\address[MU]{Electronic and Computer Science Department, Faculty of Engineering, Mondragon Unibertsitatea, 20500 Arrasate, Spain}
	\address[unitn]{Automatic Control Lab, Department of Industrial Engineering, University of Trento, Trento, Italy.}
	\address[aalto]{Intelligent Robotics Group, Department of Electrical Engineering and Automation at Aalto University, Aalto, Finland.}
	
	\begin{abstract}
		Learning from demonstration (LfD) is considered as an efficient way to transfer skills from humans to robots. Traditionally, LfD has been used to transfer Cartesian and joint positions and forces from human demonstrations. The traditional approach works well for some robotic tasks, but for many tasks of interest, it is necessary to learn skills such as orientation, impedance, and/or manipulability that have specific geometric characteristics.  An effective encoding of such skills can be only achieved if the underlying geometric structure of the skill manifold is considered and the constrains arising from this structure are fulfilled during both learning and execution.
		However, typical learned skill models such as dynamic movement primitives (DMPs) are limited to Euclidean data and fail in correctly embedding quantities with geometric constraints. In this paper, we propose a novel and mathematically principled framework that uses concepts from Riemannian geometry to allow DMPs to properly embed geometric constrains. The resulting DMP formulation can deal with data sampled from any Riemannian manifold including, but not limited to, unit quaternions and symmetric and positive definite matrices. The proposed approach has been extensively evaluated both on simulated data and real robot experiments. The performed evaluation demonstrates that beneficial properties of DMPs, such as convergence to a given goal and the possibility to change the goal during operation,  apply also to the proposed formulation. 
	\end{abstract}
	
	
\end{frontmatter}

\section{Introduction}

Reliable execution of robotic tasks in highly unstructured and dynamic scenarios is fundamental to \change{bringing} robots into human-inhabited environments. In such environments, robots need to accurately control their motion in free space as well as during physical interactions, which requires the capability to generate and adapt online reference behaviors in the form of motion, impedance, and/or force trajectories. Therefore, an effective encoding of diverse trajectory data is the key to \change{spreading} robotic solutions in everyday environments.  

The \ac{lfd} paradigm \cite{schaal1999Is} aims \change{to develop} learning solutions that allow the robot to enrich its skills via human guidance. Among the existing approaches \cite{Billard2016Learning, Ravichandar2020Recent}, the idea of encoding robotic skills into stable dynamical systems has gained interest in the \ac{lfd} community~\cite{ijspeert2013, Khansari2011learning, zadeh2014learning}. \acp{dmp} \cite{Ijspeert2002Learning} are one of the first and most popular dynamical system-based approaches for \ac{lfd}. \acp{dmp} are capable of encoding both discrete and periodic robotic skills into time-dependent systems. Discrete skills, also referred \change{to} as point-to-point motions, constist of motion trajectories with a fixed start and end point (goal) and are well-suited to represent many human daily tasks such as picking and placing objects. 

The original \ac{dmp} formulation considers one \ac{dof} trajectories. Multi-\ac{dof} trajectories are learned separately for each \ac{dof} and synchronized by a common phase variable. This strategy is effective for encoding independent skills like joint or Cartesian position trajectories, but it fails if the different \acp{dof} are mutually dependent. This situation is common in robotics, where variables of interest may be interrelated by geometric constraints. Examples of such variables include: (\emph{i}) orientation representations, like rotation matrices \cite{Ude2014} or unit quaternions \cite{Ude2014,koutras2020correct,huang2020toward}, and (\emph{ii}) inertia \cite{traversaro2016identification}, manipulability \cite{yoshikawa1985manipulability,abudakka2021probabilistic}, stiffness, and damping \cite{ikeura1995,AbuDakka2018} that are encapsulated in \ac{spd} matrices. For variables interrelated by geometric constraints, the embedding strategy has to be modified to fulfill the constraints during both training and execution. 

Several robotic skills consist of a combination of variables belonging to different manifolds. A simple example is a pose trajectory where the position lies in Cartesian space and the orientation is represented \eg as unit quaternions. 
\change{To avoid accuracy loss}, Riemannian metrics should be embedded in the \ac{dmp} formulation, \change{allowing the consideration of} all the constraints arising from various geometric structures in a unified and consistent manner\change{. This} is not possible with existing \ac{dmp} formulations~\cite{ijspeert2013,Ude2014,koutras2020correct,AbuDakka2015,abudakka2020Geometry}\change{, which} are space-dependent.  

In this paper, we propose \ac{gadmp}, a new formulation that uses differential geometry to extend classical \ac{dmp} for Euclidean data to other Riemannian manifolds. This extension allows discrete \acp{dmp} to effectively represent data evolving on different Riemannian manifolds, which subsequently allows the generation of smooth trajectories for data that do not belong to the Euclidean space. The formulation allows to encode various forms of point-to-point manipulation skills with specific geometric constraints in a unified and manifold independent manner. The general formulation provided in this paper can be applied to any trajectory of data by considering the corresponding Riemannian manifold. The effectiveness of the proposed approach is demonstrated both on synthetic data and physical experimental setups.

\begin{table*}[t]
	\centering
	\caption{Comparison among the state-of-the-art of \ac{dmp}-based approaches and our \ac{gadmp} across different Riemannian manifolds: Euclidean space of dimension $m$ $\manR{m}$, unit quaternion space $\manS{3}$, $m$-unit sphere manifold $\manS{m}$, 3D-rotation matrices space $\manSO{3}$, special orthogonal group in $m$ dimensions $\manSO{m}$, and the space of $m\times m$ SPD matrices $\spd{m}$.}
	\resizebox{\linewidth}{!}{%
		{\renewcommand\arraystretch{1}
			\begin{tabular}{m{0.35\linewidth}>{\centering\arraybackslash}m{0.05\linewidth}>{\centering\arraybackslash}m{0.05\linewidth}>{\centering\arraybackslash}m{0.05\linewidth}>{\centering\arraybackslash}m{0.05\linewidth}>{\centering\arraybackslash}m{0.05\linewidth}>{\centering\arraybackslash}m{0.05\linewidth}>{\centering\arraybackslash}m{0.16\linewidth}}
				\cline{2-8}
				& $\manR{m}$ & $\manS{3}$ & $\manS{m}$ & $\manSO{3}$ & $\manSO{m}$ & $\spd{m}$ & Composite spaces  \eg $\manS{3} \times \manR{3}$ \\ \hline
				Ijspeert~\etal~\cite{ijspeert2013,Ijspeert2002Learning}	& \checkmark & - & - & - & - & - & - \\ \hline
				Ude~\etal~\cite{Ude2014}		& - & \checkmark & - & \checkmark & - & - & - \\ \hline
				Koutras~\etal~\cite{koutras2020correct}, Abu-Dakka~\etal~\cite{AbuDakka2015},  Saveriano~\etal~\cite{saveriano2019merging}	& - & \checkmark & - & - & - & - & - \\ \hline
				Abu-Dakka~\etal~\cite{abudakka2020Geometry}	& - & - & - & - & - & \checkmark & - \\ \hline
				Our \ac{gadmp}	& \checkmark & \checkmark & \checkmark & \checkmark & \checkmark & \checkmark & \checkmark \\ \hline
			\end{tabular}
		}
	}
	\label{tab:statOfArt}
\end{table*}

Preliminary results of this work have been published in~\cite{abudakka2020Geometry}, where we formulated \ac{dmp} equations to learn \ac{spd} data profiles. This paper adds several significant novel contributions with respect to our published work:
\begin{enumerate}
	\item A unified and mathematically principled framework, \ac{gadmp}, that uses differential geometry to extend classical \acp{dmp} to any Riemannian manifold.
	\item Exploitation of manifold composites to encode and learn composite manifolds in one single \ac{dmp} formulation.
	\item Proof of the stability of the proposed \ac{gadmp}.
	\item Formulation of \ac{gadmp} goal switching without the need to use parallel transport.
	\item An extensive evaluation and comparison with existing approaches.
	\item Instructive and unified source codes accompany the paper with all necessary datasets at \url{https://gitlab.com/geometry-aware/ga-dmp}.
\end{enumerate}

This paper is organized as follows: Next section presents the state-of-the-art. A background about standard \acp{dmp} and Riemannian geometry are given in \secref{sec:background}. Afterwards, we provide the theoretical foundation of \acp{gadmp} in \secref{sec:proposed}. Subsequently, we evaluate our approach in several experiments (\secref{sec:exper}). The work is concluded in \secref{sec:concl}.

\section{Related Works}
\label{sec:relatedworks}

\ac{lfd} is a valuable framework to teach the robot new skills without explicitly coding them. \ac{lfd} framework is effective in extracting relevant patterns from a few task demonstrations and in generalizing these patterns to different scenarios. \ac{lfd} has been deeply investigated and several approaches have been developed in the literature. These include, among others, \ac{dmp} \cite{ijspeert2013, saveriano2023dynamic}, \ac{promp} \cite{paraschos2013}, \acp{gmm} \cite{calinon2014}, and \ac{kmp} \cite{huang2020toward}.   

In many previous works, training data are simply treated as time series of Euclidean vectors. Other approaches, like~\cite{pastor2009learning}, learn and adapt quaternion trajectories without enforcing the unit norm constraint, which leads to non-unit quaternions and hence requires an additional re-normalization step. Nevertheless, several works in the literature have investigated, to some \change{extent,} the problem of learning manipulation skills with specific geometric \change{constraints.} Examples of such skills include orientations, impedance, and manipulability matrices \change{that} are encapsulated in \ac{spd} matrices. The following paragraphs examine the state-of-the-art approaches.

\textbf{\ac{dmp}-based approaches:} For instance, Abu-Dakka~\etal extended the classical \acp{dmp} to encode discrete~\cite{AbuDakka2015} and periodic~\cite{abudakka2021Periodic} unit quaternion trajectories, while the work in~\cite{Ude2014} also considers different formulation to cope with rotation matrices. The quaternion-based \acp{dmp} were also extended to include the real-time goal switching mechanism~\cite{Ude2014}. The stability of the orientation \acp{dmp} is shown in~\cite{saveriano2019merging}. In~\cite{koutras2020correct}, authors proposed a modified formulation of unit quaternion \acp{dmp} to prevent oscillations that may arise in some cases. Abu-Dakka and Kyrki~\cite{abudakka2020Geometry} reformulated \acp{dmp} to generate discrete \ac{spd} profiles, which is also able to adapt to a new goal-\ac{spd}-point. There is an important conceptual difference, about how we fit a curve to data points of a demonstration on a manifold, between \ac{gadmp} and our previous work~\cite{abudakka2020Geometry}. In~\cite{abudakka2020Geometry}, to fit a curve to data points $\{\PP_t\}_{t=0}^T$ on a Riemannian manifold $\calM$, we sought a curve $\gamma:[t_0,t_T]\rightarrow\calM$ that passed exactly through each point of the demonstration trajectory. That assumption does not guarantee proximity between each pair of consecutive points, and, as detailed in Sec.~\ref{sec:dmpFormulation}, this led to the need \change{to use} parallel transport to accurately compute the \textit{covariance derivative}. However, in this paper, inspired by \cite{gousenbourger2019data}, we look for $\gamma$ to be sufficiently straight while passing sufficiently close to the data points at the given intervals. This lets us remove the parallel transport operation, i.e., to approximate the covariant derivative with the total derivative, resulting in a more compact formulation and a more efficient implementation of \ac{gadmp}.

Finally, unlike our unified formulation, the formulations of all these previously mentioned approaches are space-specific and do not consider the possibility of treating data from different manifolds in a unified and consistent manner. \Tabref{tab:statOfArt} compares our proposed \ac{gadmp} and the state-of-the-art of the \ac{dmp}-based approaches.

\textbf{Alternative approaches:}
Point-to-point motions are of particular interest in robotics as they form the basis of many everyday manipulation tasks. Therefore, researchers have developed approaches alternative to \acp{dmp} to represent point-to-point motions. Focusing on variable orientation profiles, \cite{kim2017gaussian} extended \acp{gmm} to represent the distribution of the quaternion displacements. Starting from this extended \ac{gmm}, the work in~\cite{Zeestraten2017} exploits the Riemannian structure of the unit sphere to encode variable orientations into a geometry-aware \ac{tpgmm}~\cite{calinon2014}. \ac{kmp} are extended to unit quaternions in~\cite{huang2020toward} by projecting orientation data onto the tangent space of the unit sphere (which is locally Euclidean). Learning is performed in the tangent space and generated data are projected back to the manifold.

\ac{spd} matrices are used to encapsulate data in many applications, including brain-computer interfaces~\cite{dodero2015kernel}, transfer learning~\cite{herath2017learning}, diffusion tensor imaging~\cite{alexander2001spatial}, as well as various robotic skills~\cite{Calinon2020Gaussians}. 
Alternative to \ac{dmp}, the method in~\cite{Jaquier2017} used a tensor-based formulation of \ac{gmm} and \ac{gmr} on the \ac{spd} that enabled learning and reproducing skills involving \ac{spd} without additional data re-parametrization. 
Recently, \cite{abudakka2021probabilistic} proposed a kernelized treatment to learn and adapt \ac{spd} profiles in the tangent space of the \ac{spd} manifold.

\textbf{\ac{gadmp} vs. state-of-the-art:} The aforementioned geometry-aware formulations are space-specific and do not consider the possibility of treating data from different manifolds in a unified and consistent manner. On the contrary, our \ac{gadmp} formulation is general and can be applied to any trajectory of data even when different \acp{dof} belong to different spaces.
Moreover, \acp{dmp} are one of the most popular \ac{lfd} approaches and many robotics applications rely on them. In this respect, \ac{gadmp} provides a useful framework to let users already familiar with \acp{dmp} to develop new applications. 

\section{Preliminaries}
\label{sec:background}

In this section, we briefly introduce the classical formulation of discrete \acp{dmp} (\secref{sec:dmps}) and define fundamental operations on Riemannian manifolds (\secref{sec:riemannian}). \Tabref{tab:notation} summaries the key notations used in this paper.

\begin{table*}[t]
	\small\sf\centering
	\caption{Key notations. Indices, super/subscripts, constants, and variables have the same meaning over the entire text.}
	\resizebox{\linewidth}{!}{%
	{\renewcommand\arraystretch{1} 
		
		\centering
		\begin{tabular}{m{0.15\linewidth}m{0.01\linewidth}m{0.34\linewidth}||m{0.155\linewidth}m{0.01\linewidth}m{0.335\linewidth}}
			\noalign{\hrule height 1.5pt}
			\rowcolor{Light0}
			mathcal symbols \eg $\calM$ & $\triangleq$ & denote manifolds.
			&
			bold mathcal symbols \eg $\bcalP$ & $\triangleq$ & denote trajectories. \\
			\rowcolor{Light0}
			capital letter variables \eg $\PP$ & $\triangleq$ & denote points in a manifold.
			&
			small letter variables \eg $\p$ & $\triangleq$ & denote points in a tangent space. \\
			\rowcolor{Light0}
			$\TPM$ & $\triangleq$ & The tangent space of a manifold $\calM$ around a point $\PP$
			&
			++ & $\triangleq$ & ++ \\
			\rowcolor{Light1}
			$\manR{m}$ & $\triangleq$ & Euclidean space of dimension $m$.
			&
			$\manS{m}$ & $\triangleq$ & Sphere manifold of dimension $m$. \\
			\rowcolor{Light1}
			$\manSO{m}$ & $\triangleq$ & Special orthogonal group of dimension $m$.
			&
			$\manSE{m}$ & $\triangleq$ & Special Euclidean group of dimension $m$. \\
			\rowcolor{Light1}
			$\spd{m}$ & $\triangleq$ & Space of $m \times m$ \ac{spd}.
			&
			$\sym{m}$ & $\triangleq$ & Space of $m \times m$ symmetric matrices. \\
			\rowcolor{Light2}
			$N$ & $\triangleq$ & \# of nonlinear basis functions
			&
			$i$ & $\triangleq$ & index $: i=1,2,\ldots,N$ \\
						\rowcolor{Light2}
			$l$ & $\triangleq$ & index $: l=1,2,\ldots,T$ 
			&
			$T$ & $\triangleq$ & Number of samples \\
			\rowcolor{Light2}
			$y,\dy$ & $\triangleq$ & trajectory data and its 1st derivative in classical $\ac{dmp}$
			&
			$z,\dz$ & $\triangleq$ & scaled velocity and acceleration in \ac{gadmp} \\
			\rowcolor{Light2}
			$\bcalY,\bcalYd$ & $\triangleq$ & trajectory data and its 1st derivative in \ac{gadmp}
			&
			$\bcalZ,\bcalZd$ & $\triangleq$ & scaled velocity and acceleration in \ac{gadmp} \\
			\rowcolor{Light2}
			$\alpha_z,\beta_z,\alpha_x,\alpha_g$ & $\triangleq$ & Positive constant gains.
			&
			$x$ & $\triangleq$ & \ac{dmp} phase variable. \\
			\rowcolor{Light2}
			$f(x)$, $\bm{\mathcal{F}}(x)$ & $\triangleq$ & forcing term for different spaces
			&
			$w_i$ & $\triangleq$ & adjustable weights \\
			\rowcolor{Light2}
			$\Psi_i$ & $\triangleq$ & Gaussian basis functions
			&
			$c_i$ and $h_i$ & $\triangleq$ &  centers and widths of $\Psi_i$ \\
			\rowcolor{Light2}
			$g\in \mathbb{R}$ and $\G\in\calM$ & $\triangleq$ & attractor point (goal) in different spaces
			&
			$\bcalYhat\in\calM$ & $\triangleq$ &  new manifold trajectory generated by \ac{gadmp} \\
			\noalign{\hrule height 1.5pt} 
		\end{tabular}
}}

	\label{tab:notation}
\end{table*} 

\subsection{Dynamic Movement Primitives}
\label{sec:dmps}

\ac{dmp} is composed of a system of nonlinear differential equations capable of encoding movements while guaranteeing convergence to a designated goal point (attractor) \cite{saveriano2023dynamic}. The foundational work on \acp{dmp} for discrete, point-to-point, motions was first introduced by Ijspeert~\etal~\cite{Ijspeert2002Learning}. However, in order to generate movements adaptable to new situations without inducing excessive accelerations or amplification, Pastor~\etal introduced some modifications \cite{pastor2009learning}. In this paper, we adopt the formulation proposed by Pastor~\etal.
For a single \ac{dof} trajectory $y$, the \ac{dmp} system of equations proposed in \cite{pastor2009learning} is described as follows:
\begin{align}
	\tau\dot{z} & = \alpha_z(\beta_z(g - y - (g-y_0)x + f(x))-z), \label{eq:DMP_accel} \\
	\tau\dot{y} & = z, \label{eq:DMP_velocity}\\
	\tau\dot{x} & = -\alpha_xx, \label{eq:DMP_phase}
\end{align}
where $\tau$ is a positive scalar \change{that} represents the temporal scaling factor and determines the overall duration of the movement. $\dot{y}$ represents velocity and $z$ denotes scaled velocity. $x$ is a phase variable, governing the dynamical system's evolution towards the attractor point. It is used to avoid explicit time dependency in the formulation. The canonical system, given by \eqref{eq:DMP_phase}, is initialized as $x(0)=1$ and vanishes exponentially\footnote{The minimum phase to execute a motion within $T_f$ seconds can be computed through $x(T_f)=\text{exp}(-\frac{\alpha_x}{\tau}T_f)$.} as $t\rightarrow\infty$ if the gain $\alpha_x>0$. 
$\beta_z$ and $\alpha_z$ are positive gains that define the dynamical system's behavior. In order to ensure a critically damped system, we choose $ \alpha_z= 4{\beta_z}$. The attractor (goal) point of the movement is denoted by $g$. This system of equations prevents high accelerations at the beginning of the motion or when the goal is close to the initial state, allowing for the reproduction of motions with the same initial and target states while preventing over-amplifications and trajectory mirroring effects when changing the goal.

The nonlinear forcing term $f(x)$ is classically parameterized as a linear combination of $N$ nonlinear radial basis functions scaled by the phase variable $x$. $f(x)$ allows the dynamical system to preserve the shape of any smooth trajectory, and subsequently, generate this trajectory from an initial position $y_0$ to the attractor $g$. 
Thus, $f(x)$ is defined as:
\begin{equation}
	f(x) = \frac{\sum_{i=1}^Nw_i\Psi_i(x)}{\sum_{i=1}^N\Psi_i(x)}x,
	\label{eq:dmp_fx}
\end{equation}
\begin{equation}
	\Psi_i(x) = \exp\left(-h_i\left(x-c_i\right)^2\right),
	\label{eq:dmp_psi}
\end{equation}
where $w_i$ are the weights adjusted based on measured data to achieve the desired behavior. $\Psi_i(x)$ are Gaussian basis functions with centers $c_i$ and widths $h_i$. For a given number of basis functions $N$, centers $c_i$ and widths $h_i$ are defined as follows: 
\begin{equation*}
    c_i=\text{exp}\big(-\alpha_x\frac{i-1}{N-1}\big), \,\,
    h_i=\frac{1}{(c_{i+1} - c_i)^2}, \,\,
    h_N=h_{N-1}
\end{equation*}
where $i = 1, \dots, N$. 
For each \ac{dof}.

In order to control multiple \acp{dof} systems, such as trajectories of joint angles of $D$ \ac{dof} manipulator, we consider a separate transformation system \eqref{eq:DMP_accel}-\eqref{eq:DMP_velocity} for each of the $D$ \acp{dof} to control. Additionally, we utilize a single canonical system \eqref{eq:DMP_phase} shared across the $D$ transformation systems, which synchronizes their time evolution.

\subsection{Riemannian manifolds}
\label{sec:riemannian}

An $m$-dimensional manifold is a topological space where each point locally resembles Euclidean space $\manR{m}$. A differentiable manifold extends this notion to ensure that at each point, there exists a tangent space. A Riemannian manifold $\calM$ is a smooth and differentiable manifold where each tangent space is equipped with a Riemannian metric tensor. This tensor, denoted as $\langle\cdot,\cdot\rangle_\PP$, is a positive definite inner product defined on the tangent space ${\TPM}$ for every point ${\PP\in\calM}$. The Riemannian metric introduces the concept of length on the manifold.
By utilizing this metric, we can generalize the notion of a ``straight line'' between two points by defining a geodesic as the shortest curve that connects two points on a manifold.
This geodesic allows for the transportation of vectors between tangent spaces \cite{pennec2006intrinsic,absil2009optimization}.
A geodesic $\gamma(t)$ is defined as a continuously differentiable curve that connects points $\bm{A},\bm{B}$ on the manifold $\calM$. It locally minimizes \change{the} distance between these points, and its length is given by the functional:
\begin{equation}
	\bm{\calL}_{\bm{A}}^{\bm{B}}(\gamma) = \int_{0}^{1} \langle\dot{\gamma}(t),\dot{\gamma}(t)\rangle \, \text{d}t.
	\label{eq:geodesic}
\end{equation}
The distance between points $\bm{A}$ and $\bm{B}$ is then defined by minimizing \eqref{eq:geodesic}, \ie
\begin{equation}
	\text{dist}(\bm{A},\bm{B}) = \text{min}\,\bm{\calL}_{\bm{A}}^{\bm{B}}(\gamma)
	\label{eq:Riemannian_distance}
\end{equation}

\begin{figure}[b]
	\centering
	\def\svgwidth{\linewidth}
	{\fontsize{9}{9}
\begingroup%
  \makeatletter%
  \providecommand\color[2][]{%
    \errmessage{(Inkscape) Color is used for the text in Inkscape, but the package 'color.sty' is not loaded}%
    \renewcommand\color[2][]{}%
  }%
  \providecommand\transparent[1]{%
    \errmessage{(Inkscape) Transparency is used (non-zero) for the text in Inkscape, but the package 'transparent.sty' is not loaded}%
    \renewcommand\transparent[1]{}%
  }%
  \providecommand\rotatebox[2]{#2}%
  \newcommand*\fsize{\dimexpr\f@size pt\relax}%
  \newcommand*\lineheight[1]{\fontsize{\fsize}{#1\fsize}\selectfont}%
  \ifx\svgwidth\undefined%
    \setlength{\unitlength}{235.81182285bp}%
    \ifx\svgscale\undefined%
      \relax%
    \else%
      \setlength{\unitlength}{\unitlength * \real{\svgscale}}%
    \fi%
  \else%
    \setlength{\unitlength}{\svgwidth}%
  \fi%
  \global\let\svgwidth\undefined%
  \global\let\svgscale\undefined%
  \makeatother%
  \begin{picture}(1,0.49375551)%
    \lineheight{1}%
    \setlength\tabcolsep{0pt}%
    \put(0,0){\includegraphics[width=\unitlength,page=1]{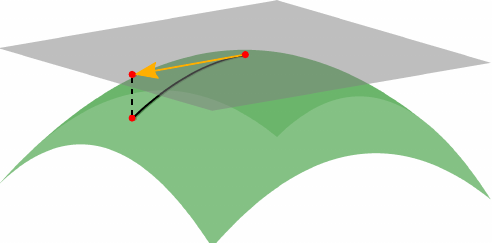}}%
    \put(0.42102554,0.05865716){\color[rgb]{0,0,0}\rotatebox{-0.9952857}{\makebox(0,0)[lt]{\lineheight{1.25}\smash{\begin{tabular}[t]{l}$\calM$\end{tabular}}}}}%
    \put(0.48407173,0.40730274){\color[rgb]{0,0,0}\rotatebox{-0.9952857}{\makebox(0,0)[lt]{\lineheight{1.25}\smash{\begin{tabular}[t]{l}$\bm{P}$\end{tabular}}}}}%
    \put(0.76475133,0.33885067){\color[rgb]{0,0,0}\rotatebox{8.82827481}{\makebox(0,0)[lt]{\lineheight{1.25}\smash{\begin{tabular}[t]{l}$\TPM$\end{tabular}}}}}%
    \put(0.3385989,0.24581842){\color[rgb]{0,0,0}\rotatebox{-0.9952857}{\makebox(0,0)[lt]{\lineheight{1.25}\smash{\begin{tabular}[t]{l}$\aEXPmap$\end{tabular}}}}}%
    \put(0.00891622,0.32658385){\color[rgb]{0,0,0}\rotatebox{-0.9952857}{\makebox(0,0)[lt]{\lineheight{1.25}\smash{\begin{tabular}[t]{l}$\aLOGmap$\end{tabular}}}}}%
    \put(0,0){\includegraphics[width=\unitlength,page=2]{ts.pdf}}%
  \end{picture}%
\endgroup%
}
	\caption{A Riemannian manifold $\calM$ and its tangent space $\TPM$ defined at point $\bm{P}$.}
	\label{fig:ts}
\end{figure}

\subsubsection{Mapping operators}
The tangent spaces and their bases provide the ability to perform linear algebra. In order to perform computations on the manifold while preserving distances, a mapping system is needed to switch between the tangent space $\TPM$ and the manifold $\calM$, see \figref{fig:ts}. These mapping operators are:
\begin{itemize}[leftmargin=10pt,noitemsep]
	\item \textbf{The logarithmic map} $\left(\Log{\PP}{\cdot}\right)$ is a function that maps a point ${\bm{A}\in\calM}$ into a point $\bm{a}\in\TPM$ (see \figref{fig:ts}). It is defined as:
	\begin{equation}
	    \Log{\PP}{\cdot}\!:\!\calM\mapsto\TPM, \label{eq:LOG}
	\end{equation}
    \item \textbf{The exponential map} $\left(\Exp{\PP}{\cdot}\right)$ is the inverse of the logarithmic map. It maps a point $\bm{a}\in\TPM$ in the tangent space of $\bm{P}$ to a point ${\bm{A}\in\calM}$ such that $\bm{A}$ lies on the geodesic starting from $\PP$ in the direction of $\bm{a}$ with distance of  $\norm{\bm{a}}=\langle\bm{a},\bm{a}\rangle_\PP$ (see \figref{fig:ts}). It is defined as:
	\begin{equation}
	    \Exp{\PP}{\cdot}\!:\!{\TPM\mapsto\calM}, \label{eq:EXP}
	\end{equation}
\end{itemize}

\subsubsection{Cartesian products in Riemannian geometry}

In Riemannian geometry, the Cartesian product of two Riemannian manifolds $\calM$ and $\calN$ is also a manifold denoted as $\calM \times \calN$. This construction allows us to combine the geometric structures of both $\calM$ and $\calN$ into a single manifold.

For any points $\PP_1\in \calM$ and $\U_1\in \calN$, and their corresponding tangent vectors $\bm{p}_1 \in \TPPM$ and $\bm{u}_1 \in \TUUN$, the tangent space of $\calM \times \calN$ at the point $(\PP_1,\U_1)$ is isomorphic to the direct sum of the tangent spaces of $\calM$ and $\calN$:

\begin{equation}
	\calT_{(\PP_1,\U_1)}(\calM \times \calN) \cong \TPPM \oplus \TUUN, \label{eq:cartProd}
\end{equation}

This means that any tangent vector at $(\PP_1,\U_1)$ can be uniquely decomposed into a pair of tangent vectors, one in $\TPPM$ and the other in $\TUUN$.

To facilitate computations on the Cartesian product manifold $\calM \times \calN$, we require to redefine the mapping operators in \eqref{eq:LOG} and \eqref{eq:EXP} as follows: 
\begin{align}
	\Log{(\PP_1,\U_1)}{\PP_2,\U_2} &: \calM \times \calN \mapsto \calT_{(\PP_1,\U_1)}(\calM \times \calN), \label{eq:logCP} \\
	\Exp{(\PP_1,\U_1)}{\bm{p},\bm{u}} &:\calT_{(\PP_1,\U_1)}(\calM \times \calN) \mapsto \calM \times \calN. \label{eq:expCP}
\end{align}
This leads to 
\begin{align*}
	\Log{(\PP_1,\U_1)}{\PP_2,\U_2} &= \Log{\begin{bmatrix}
			\PP_1 \\ \U_1
	\end{bmatrix}}{\begin{bmatrix}
			\PP_2 \\ \U_2
	\end{bmatrix}} = \begin{bmatrix}
		\Log{\PP_1}{\PP_2} \\ \Log{\U_1}{\U_2}
	\end{bmatrix}, \\ 
	\Exp{(\PP_1,\U_1)}{\bm{p},\bm{u}} &= \Exp{\begin{bmatrix}
			\PP_1 \\ \U_1
	\end{bmatrix}}{\begin{bmatrix}
			\bm{p} \\ \bm{u}
	\end{bmatrix}} = \begin{bmatrix}
		\Exp{\PP_1}{\PP_2} \\ \Exp{\U_1}{\U_2}
	\end{bmatrix}.
\end{align*}
where $(\bm{p},\bm{u}) \in \calT_{(\PP_1,\U_1)}(\calM \times \calN)$ and  $(\PP_2,\U_2) \in \calM \times \calN$.

\subsubsection{Computing in Riemannian manifolds}
Let $\PP_1,\PP_2\in \calM$ and $\p_1,\p_2\in \manR{m}$, then the reinterpretation of basic standard operations (\eg addition and subtraction) in a Riemannian manifold are summarized in \tabref{tab:basicOperation}.

\begin{table}[ht]
	\small\sf
	\centering
	\caption{Re-interpretation of basic standard operations in a Riemannian manifold \cite{Pennec2006}.}
	\resizebox{\linewidth}{!}{%
	{\renewcommand\arraystretch{1.2} 
		\begin{tabular}{m{0.17\linewidth}m{0.4\linewidth}m{0.42\linewidth}}
			\Cline{3pt}{2-3}
				& Euclidean space 	& Riemannian manifold		\\
				\noalign{\hrule height 1.5pt}
			Subtraction	& $\overrightarrow{\p_1\p_2}=\p_2-\p_1$	& $\overrightarrow{\PP_1\PP_2}=\Log{\PP_1}{\PP_2}$		\\
			Addition	& $\p_2 = \p_1+\overrightarrow{\p_1\p_2}$ & $\PP_2=\Exp{\PP_1}{\overrightarrow{\PP_1\PP_2}}$		\\
			Distance	& $\text{dist}(\p_1,\p_2)=\parallel \p_2-\p_1\parallel$ & $\text{dist}(\PP_1,\PP_2)=\parallel\overrightarrow{\PP_1\PP_2}\parallel_{\PP_1}$		\\
			Interpolation	& $\p(t)=\p_1+t \overrightarrow{\p_1\p_2}$ & $\PP(t)=\Exp{\PP_1}{t\overrightarrow{\PP_1\PP_2}}$		\\
			\noalign{\hrule height 1.5pt}
		\end{tabular}
    }}
	\label{tab:basicOperation}
\end{table} 

\subsubsection{Riemannian geometric mean} 
Given a set of points $\left\{\PP_i\right\}_{i=1}^n\in\mathcal{M}$ and a geodesic distance $\mathrm{dist}(\PP_j,\PP_i)$ between two points in $\mathcal{M}$, the Fréchet mean~\cite{frechet1948elements} is estimated by minimizing the sum of squared geodesic distances
\begin{equation}
    \overline{\PP}\mathop = \arg\min_{\PP\in{\mathcal{M}}}\sum_{i=1}^N \mathrm{dist}^2(\PP,\PP_i),
    \label{eq:mean}
\end{equation}
This estimation can be efficiently computed iteratively by following Alg.~\ref{alg:mean}~\cite{frechet1948elements}.
\begin{algorithm}[h]
    \textbf{Initialization:} $\overline{\PP} = \PP_1$
    \begin{algorithmic}[1]
    \While{\texttt{$\norm{\bm{a}}<\delta$}}
        \State{$\bm{a}=\frac{1} {N}\sum_{i=1}^N\Log{\overline{\PP}}{\PP_i}$}       \State{$\overline{\PP}=\Exp{\overline{\PP}}{\epsilon\bm{a}};\,\, \epsilon\leq1$}
    \EndWhile
    \end{algorithmic}
    \caption{Intrinsic mean}
    \label{alg:mean}
\end{algorithm}

\section{Proposed approach}
\label{sec:proposed}
In this section, we provide a generalized and unified formulation for \acp{dmp} based on Riemannian geometry in order to learn and adapt robot manipulation skills regardless its corresponding space, for example orientation trajectories $\left(\manSO{3} \text{ or } \manS{3}\right)$, pose data ($\manSE{3}$), and \ac{spd} profiles ($\spd{m}$) such as stiffness, manipulability, inertia. We also show that our \ac{gadmp} inherits desirable properties of the original formulation like convergence to a target and goal switching.

\subsection{Geometry-aware \texorpdfstring{\acp{dmp}}{} formulation}
\label{sec:dmpFormulation}

In this section, we introduce the mathematical foundations of \ac{gadmp} technique. The \ac{gadmp} formulation offers a comprehensive and cohesive approach to encode and execute a discrete trajectory ${\bcalY = \{t_l, \Y_l\}_{l=0}^T}$, commonly known as a point-to-point trajectory, which evolves within the confines of a Riemannian manifold $\calM$, where each $\Y_l \in \calM$. 
Its attractor dynamics on the manifold guarantee the convergence of $\bcalY$ toward a goal $\G\in\calM$ regardless of the initial starting point $\Y_0$. To achieve this, it is necessary to transform the classical \ac{dmp} system described by \eqref{eq:DMP_accel}--\eqref{eq:DMP_velocity} into a unified geometry-aware formulation utilizing principles from Riemannian geometry. In pursuit of this objective, we initiate the process by considering the expression of a general second-order system evolving on a manifold, as outlined by Fiori \etal\cite{fiori2022synthetic}

\begin{align}
    \tau\nabla_{\bcalZ}\bcalZ &= \bm{h}\left(\bcalZ, \bcalY, x\right), \label{eq:ds_manifold_1}\\
    \tau\bm{\dot{\bcalY}} &= \bcalZ,
    \label{eq:ds_manifold_2}
\end{align}
where $\bcalZ$ and $\bcalZd$ represent the scaled first and second derivatives of $\bcalY$. The phase variable $x$ is similar to the one defined in \eqref{eq:DMP_accel} and \eqref{eq:DMP_phase}.  The \textit{covariant derivative} $\nabla_{\bcalZ}\bcalZ$ can be defined from the total derivative $\bcalZd$ using parallel transport~\cite{fiori2022synthetic,abudakka2020Geometry}. However, computing the parallel transport is, in general, time-consuming. Assuming that consecutive points on the manifold are sufficiently close, and the geodesic between them approximates a straight line, the covariant derivative can be well approximated by manifold-valued finite differences~\cite{boumal2011discrete, gousenbourger2019data}. This approximation significantly simplifies the computation process while introducing negligible errors. Thus, in this work, we consider the approximation $\nabla_{\bcalZ}\bcalZ \approx \bcalZd$. The function $\bm{h}(\cdot)$ may encompass multiple additive contributions. In this study, we assume that
\begin{align}
    \bm{h}\left(\bcalZ, \bcalY, x\right) &= \alpha_z\left(\beta_z\left(\Log{\bcalY}{\G} \right.\right. \nonumber \\
    &- \left.\left.\Log{\Y_0}{\G} x + \bcalF(x) \right)- \bcalZ\right),
\end{align}
where $\bm{G} \in \mathcal{M}$ is the goal point. The function $\Log{\bcalY}{\cdot}$ is defined in~\eqref{eq:LOG}. Additionally, positive gains $\alpha_z$ and $\beta_z$ are introduced. The term $-\alpha_z\bcalZ$ represents a \textit{dissipative force} that plays a similar role to damping in a mechanical system. The term $\alpha_z(\beta_z \, \Log{\bcalY}{\G})$ corresponds to \textit{conservative force} and can be interpreted as the negative gradient of a potential. This can be demonstrated by considering that ${-\frac{1}{2}\nabla_{\bcalY}\mathrm{dist}^2\left(\bcalY,\bm{G}\right) = \Log{\bcalY}{\G}}$~\cite{fiori2022synthetic}, where $\mathrm{dist}(\cdot,\cdot)$ denotes the Riemannian distance. Finally, the term $\bcalF(x)$ represents a phase-dependent forcing term which is learned from the demonstration and will be further discussed in this section.

Consequently, we can redefine the dynamic system presented in \eqref{eq:DMP_accel}--\eqref{eq:DMP_velocity} as follows
\begin{align}
    \tau\bcalZd & =  \alpha_z\left(\beta_z \left(\Log{\bcalY}{\G} \right.\right. \nonumber \\
    &- \left.\left.\Log{\Y_0}{\G} x + \bcalF(x) \right)- \bcalZ\right). \label{eq:DMP_RM_accel}\\
    \tau\bcalYd & =  \bcalZ, \label{eq:DMP_RM_velocity}
\end{align}
The forcing term $\bcalF(x)$ is defined as follows 
\begin{align}
		\bm{\bcalF}(x) = \frac{\sum_{i=1}^N\w_i\Psi_i(x)}{\sum_{i=1}^N\Psi_i(x)}x ,
	\label{eq:dmp_rm_fx}
\end{align}
where $\w_i\in\manR{m\times N}$ are the weights (free parameters) \change{that} can be estimated by encoding any sampled trajectory (\eg any robot manipulation skill profile). In order to estimate the parameters of a corresponding \acp{gadmp}, we need to estimate the \ordinal{1}{st} and \ordinal{2}{nd} time derivatives of the demonstrated trajectory. The 1st time derivative is computed as follows
\begin{equation}
	\bcalYd = \left\{\left(\Log{\Y_{l-1}}{\Y_{l}}\right)/\delta t\right\}_{l=1}^T ~ \in \TSM{\Y_{l-1}},
\end{equation}
where $\delta t = t_l - t_{l-1}$.
The \ordinal{2}{nd}-time-derivative $\bcalYdd$ can be computed straight forward from $\bcalYd$ using standard Euclidean tools, \ie  ${\bcalYdd = \{t_l, \ydd_l\}_{l=1}^T}$ where $\ydd_l = (\yd_{l}-\yd_{l-1})/\delta t$. 

Having all necessary data $\left\{t_l,\Y_l,\yd_l,\ydd_l\right\}$, 
and by inverting \eqref{eq:DMP_RM_accel}, the parameters $\w_i$ and the approximated desired shape of the demonstration \change{are} estimated as follows
\begin{equation}
	\begin{split}
	    &\frac{\sum_{i=1}^N\w_i\Psi_i(x_l)}{\sum_{i=1}^N\Psi_i(x_l)}x_l = \\ &
	    \frac{\tau^2\ydd_l + \alpha_z\tau\yd_l}{\alpha_z\beta_z} - \Log{\Y_l}{\G} + \Log{\Y_0}{\G} x
	\end{split}
	\label{eq:dmp_rm_ft}
\end{equation}
 Using \eqref{eq:dmp_rm_ft}, the weights $\w_i$ can be estimated by encoding any sampled robot manipulation skill data.

In the reproduction, equation \eqref{eq:DMP_RM_velocity} is integrated using the forward Euler-Riemann stepping method~\cite{fiori2022synthetic} as
\begin{equation}
	\bcalYhat(t+\delta t)=\Exp{\Y_t}{\bcalZ(t)\frac{\delta t}{\tau}},
	\label{eq:manExp}
\end{equation}
where $\bcalYhat\in \calM$ represents the new robot manipulation skills data. Equation \eqref{eq:manExp} is manifold dependent. $\Exp{\Y_t}{\cdot}$ is defined as in \eqref{eq:EXP}, and we refer to the appendix for the expression of $\Exp{\Y_t}{\cdot}$ for the manifolds used in this work.

In case the manifold is a Lie group, the expression of a general second-order system on a Lie group becomes~\cite{fiori2022synthetic}
\begin{align}
\tau\bcalZd &= \bm{h}\left(\bcalZ, \bcalY, x\right), \label{eq:ds_group_1}\\
\tau\bcalYd &= \bm{g}\left(\bcalZ, \bcalY\right), \label{eq:ds_group_2}
\end{align}
from which is straightforward to derive that
\begin{align}
\tau\bcalZd &= \alpha_z\left(\beta_z \left(\, \mathrm{Log}\left(\Y_g*\bcalY^{-1}\right) \right.\right. \nonumber \\
&- \left.\left.\mathrm{Log}\left(\Y_g*\Y_0^{-1}\right) + \bcalF(x)\right)-\bcalZ\right) , \label{eq:lie-dmp1}\\
\tau\bcalYd &= \bm{g}\left(\bcalZ, \bcalY\right). \label{eq:lie-dmp2}
\end{align}
Equation~\eqref{eq:lie-dmp1} is formally the same as~\eqref{eq:DMP_RM_accel}, provided we use the logarithmic map $\Log{\bcalY}{\cdot}=\mathrm{Log}\left(\Y_g*\bcalY^{-1}\right)$ defined using Lie group theory. The term $\bm{m}(\cdot)$ in~\eqref{eq:lie-dmp2} is the \textit{inverse left translation}, which maps a tangent vector from the Lie algebra to the tangent space at $\Y_t$ and depends on the specific Lie group. The expressions of $\bm{g}(\cdot)$ and $\mathrm{Log}(\cdot)$ for unit quaternions and rotation matrices, two Lie groups commonly used in robotics, are given in~\ref{sec:manifoldS3} and \ref{sec:manifoldSO3}.

As a remark, we used the Riemannian formulation~\eqref{eq:DMP_RM_accel}--\eqref{eq:DMP_RM_velocity} in the rest of the paper. However, for the sake of completeness, we also have provided a formulation for Lie groups in~\eqref{eq:lie-dmp1}--\eqref{eq:lie-dmp2}.

\subsection{Goal switching}
\label{sec:goalSwitching}

In many real scenarios, while the robot \change{executes} its trajectory, it may encounter situations where it needs to adapt its trajectory to a new goal, \eg new pick-up point, on the fly. This change of goal, referred to as goal switching, is a common requirement in dynamic environments. In order to achieve smooth transitions between goals and avoid unnecessary jumps, \change{the authors of} \cite{ijspeert2013} suggested \change{adding} an extra first-order differential equation to gradually transition the current goal $g$ to the new goal $g_{\text{new}}$ over time. This differential equation can be written as 
\begin{equation}
	\tau \dot{g} = \alpha_g(g_{new}-g),
	\label{eq:classic_goal_switch}
\end{equation}
where $\alpha_g > 0$ is a positive constant gain. The gradual transition in~\eqref{eq:classic_goal_switch} ensures that the robot's behavior remains continuous and responsive to changes in its task environment.

Analogously, Riemannian manifold-based \ac{gadmp} can switch the goal using
\begin{equation}
	\tau \bcalGd = \alpha_g \Log{\G}{\G_{new}}.
	\label{eq:riem_goal_switch}
\end{equation}
Equation~\eqref{eq:riem_goal_switch} allows to continuously update $\G$ until it smoothly reaches the new value $\G_{new}\in\calM$. 

\subsection{Stability analysis}
\label{sec:stability}
Theorem~\ref{th:stab_rdmp} states the stability conditions of the geometry-aware \ac{dmp} formulation in \secref{sec:dmpFormulation}.
\begin{theorem}
\label{th:stab_rdmp}
Assume that $ \bm{\bcalF}(x) \rightarrow 0$ for $t \rightarrow +\infty$ and that the gains $\alpha_z,\,\beta_z > 0$. Under these assumptions, the geometry-aware \ac{dmp} has a globally (in its domain of definition) asymptotically stable equilibrium at $(\G,\bm{0})$.
\end{theorem}
\begin{proof}
Recall that, by assumption, we restrict the domain to the points where the logarithmic map $\Log{\bcalY}{\G}$ is uniquely defined. Recall also that the forcing term $\bcalF(x)$ in~\eqref{eq:DMP_RM_accel} is a weighted sum of Gaussian basis functions. Therefore, the non-linear terms in~\eqref{eq:DMP_RM_accel}~and~\eqref{eq:DMP_RM_velocity} are smooth and uniquely defined functions. Consider also that the time \change{dependency} introduced by $x$ vanishes for $t\rightarrow +\infty$. Hence, \eqref{eq:DMP_RM_accel}~and~\eqref{eq:DMP_RM_velocity} are an \textit{asymptotically autonomous differential system} and the stability can be proved \change{by} analyzing its asymptotic behavior~\cite{Markus56}. This allows \change{us} to neglect the terms $\bcalF(x)$ and $\Log{\Y_0}{\G} x$ in the stability analysis and to focus on the asymptotic dynamics
\begin{align}
	\begin{split}
		\bcalZd & =  \alpha_z\beta_z \Log{\bcalY}{\G} - \alpha_z\bcalZ, \label{eq:DMP_RM_accel_asym}
	\end{split}\\
	\bcalYd & =  \bcalZ, \label{eq:DMP_RM_velocity_asym}
\end{align}
where we set $\tau=1$ without loss of generality. 

We first show that $(\G,\bm{0})$ is an equilibrium point of the system~\eqref{eq:DMP_RM_accel_asym}~and~\eqref{eq:DMP_RM_velocity_asym}. The right side of~\eqref{eq:DMP_RM_velocity_asym} vanishes only for $\Z = \bm{0}$. With $\Z = \bm{0}$, the right side of~\eqref{eq:DMP_RM_accel_asym} vanishes only for $\Log{\Y}{\G}=\bm{0} \Leftrightarrow \Y = \G$. This implies that the system~\eqref{eq:DMP_RM_accel_asym}~and~\eqref{eq:DMP_RM_velocity_asym} has a unique equilibrium point at $(\G,\bm{0})$.

We now show that the equilibrium $(\G,\bm{0})$ is a global attractor in the chart where the logarithmic map $\Log{\bcalY}{\G}$ is uniquely defined. To this end, we define the candidate Lyapunov function
\begin{equation}
    V(\bcalY,\bcalZ) = \text{dist}^2\left(\bcalY, \G \right) + \frac{1}{\alpha_z\beta_z} \langle \bcalZ, \bcalZ \rangle_{\bcalY},
    \label{eq:lyapunov_candidate}
\end{equation}
where $\text{dist}(\cdot,\cdot)$ is the Riemannian distance defined as in~\eqref{eq:Riemannian_distance} and $\langle \cdot, \cdot \rangle_{\bcalY}$ is the positive definite inner product (see \secref{sec:riemannian}). $V(\bcalY,\bcalZ)$ is positive definite everywhere if $\alpha_z\beta_z > 0$ and vanishes only at $\Y = \G$ ($\text{dist}^2\left(\G, \G \right) = 0$) and $\Z = \bm{0}$ ($\langle \bm{0}, \bm{0} \rangle_{\bcalY} = 0$). To show that $V(\bcalY,\bcalZ)$ is a valid Lyapunov function we need to show that its time derivative is negative definite and vanishes at $(\G,\bm{0})$. The time derivative of $V(\bcalY,\bcalZ)$ can be written as
\begin{equation}
\begin{split}
\dot{V}(\bcalY,\bcalZ) &= \frac{d}{dt}\text{dist}^2\left(\bcalY, \G \right) + \frac{1}{\alpha_z\beta_z}\frac{d}{dt}  \langle \bcalZ, \bcalZ \rangle_{\bcalY} \\
&= -2 \langle \Log{\bcalY}{\G}, \dot{\bcalY} \rangle_{\bcalY} + \frac{2}{\alpha_z\beta_z} \langle \dot{\bcalZ}, \bcalZ \rangle_{\bcalY}
\end{split}
\label{eq:lyapunov_derivative}
\end{equation}
where we used the expression $\frac{d}{dt}\text{dist}^2\left(\bcalY, \G \right) = -2 \langle \Log{\bcalY}{\G}, \dot{\bcalY} \rangle_{\bcalY}$ from~\cite{fiori2021manifold} and the bi-linearity and the symmetry of the interior product to write $\frac{d}{dt}  \langle \bcalZ, \bcalZ \rangle_{\bcalY} = 2 \langle \dot{\bcalZ}, \bcalZ \rangle_{\bcalY}$. By replacing $\dot{\bcalZ}$ from~\eqref{eq:DMP_RM_accel_asym} and $\dot{\bcalY}$ from~\eqref{eq:DMP_RM_velocity_asym} into~\eqref{eq:lyapunov_derivative}, we obtain
\begin{equation*}
\begin{split}
\dot{V}(\bcalY,\bcalZ) &= -2 \langle \Log{\bcalY}{\G}, \bcalZ \rangle_{\bcalY} + 2 \langle \Log{\bcalY}{\G}, \bcalZ \rangle_{\bcalY} \\
& - \frac{2}{\beta_z} \langle \bcalZ, \bcalZ \rangle_{\bcalY} = - \frac{2}{\beta_z} \langle \bcalZ, \bcalZ \rangle_{\bcalY} \leq 0,
\end{split}
\end{equation*}
where the last inequality holds if $\beta_z > 0$. Therefore, $\dot{V}(\bcalY,\bcalZ) \leq 0$ everywhere in the chart and vanishes only at $\Z = \bm{0}$. The LaSalle’s invariance theorem~\cite{slotine1991applied} allows to conclude the stability of \eqref{eq:DMP_RM_accel_asym}--\eqref{eq:DMP_RM_velocity_asym}.
\end{proof}

\begin{remark}
\label{rem:single_chart}
The results of Theorem~\ref{th:stab_rdmp} hold where the logarithmic map is uniquely defined, \eg $\TSM{\Y_{l-1}}$ can be extended as much as it will not contain points conjugate to $\Y_{l-1}$ \cite{pait2010properties}. For manifolds with no cut-locus, this holds everywhere. Hence, Theorem~\ref{th:stab_rdmp} is globally valid on manifolds with no cut-locus (\eg the manifold of \ac{spd} matrices with positive definite eigenvalues \cite{Pennec2006}).
However, for manifolds with cut-locus (\eg unit $m$-sphere manifolds~\cite{pennec2006intrinsic}), the logarithmic map $\Log{\bcalY}{\G}$  is defined in a region that does not contain points conjugate to $\G$. For the unit m-sphere, the logarithmic map $\Log{\bcalY}{\G}$ is uniquely defined everywhere apart from the antipodal point $-\G$.

\end{remark}
\begin{figure}[t]
	\centering
	\def\svgwidth{\linewidth}
	{\fontsize{8}{8}
		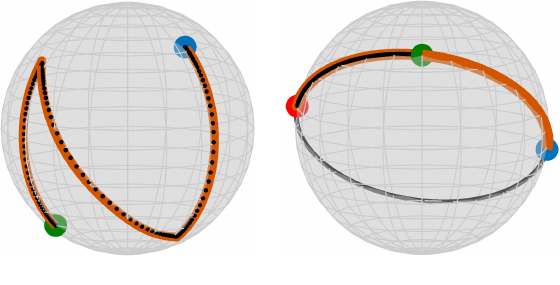}
	\caption{Results of \ac{gadmp} while learning and producing trajectories that cover both south and north hemispheres. Black dashed curves denote demonstrations, while brown curves represent reproduction. Green point $\Y_1$ denotes the starting point of the trajectory, while the blue one indicates the goal $\G$. The red point illustrates the antipodal point of the goal. The figure shows \ac{gadmp} while executing a trajectory that (a) does not contain an antipodal of the goal $\G$, \change{and} (b) contains an antipodal of the goal.}
	\label{fig:antipodal}
\end{figure}
For illustration, we used the proposed \ac{gadmp} to learn two trajectories; (\emph{i}) the ``N'' shape on $\manS{2}$ provided in \cite{Calinon2020Gaussians} (\figref{fig:antipodal}a), and (\emph{ii}) a ``C'' curve with $\pi$ diameter (\figref{fig:antipodal}b). The ``N'' trajectory covers both the north and south \change{hemispheres} and, as shown in~\cite{Calinon2020Gaussians}, working on the Lie algebra will introduce large distortions. Moreover, the ``N'' shape is an antipodal free trajectory, such that $\bcalY = \{\Y_l\}_{l=1}^{T-1} \in \manS{2} \ \vert\ \lvert{\Y_l\cdot\G}\rvert < 1$. However, the ``C'' curve includes the antipodal of $\G$. \Figref{fig:antipodal}a shows \ac{gadmp} successfully reproducing the shape and converges to the goal (blue point). However, in (b), it fails to follow the trajectory when it encounters the antipodal of the goal (point in red). \ac{gadmp} \change{is supposed} to follow the trajectory in the direction of the black arrow starting from the green point. However, it follows the trajectory until the antipodal point, then returns back to reach the goal from the opposite direction. A possible way to solve this issue is to split the trajectory into segments. For the example in \figref{fig:antipodal}b, this can be done by splitting the trajectory into $2$ segments, namely $\Y_1$ to $\Y_2$, and $\Y_2$ to $\G$, where $\Y_2$ is any point in the demonstration between $-\G$ and $-\Y_1$. One can then fit $2$ separate \ac{gadmp} and smoothly merge them~\cite{saveriano2019merging}.

\subsection{\texorpdfstring{\ac{gadmp}}{} on Riemannian manifold products}
\label{sec:compositeDMP}

Let us define $\bcalY \in \calM$ and $\bcalU \in \calN$ as two arbitrary trajectories from two Riemannian manifolds $\calM$ and $\calN$, respectively. Let us call  $\bcalH = \{t_l, (\Y_l,\U_l)\}_{l=1}^T$ the set of data points in one demonstration. We can now define the \textit{composite} \ac{gadmp} as
\begin{align}
	\begin{split}
		\tau\bcalVd & =  \alpha_z(\beta_z \Log{(\bcalY, \bcalU)}{\G_\Y,\G_\U} - \bcalV) + \bm{\bcalF}(x), \label{eq:DMP_Comp_accel}
	\end{split}\\
	\tau\bcalHd & =  \bcalV, \label{eq:DMP_Comp_velocity}
\end{align}
where $\bcalV \in \calT_{(\Y_l,\U_l)}(\calM\times \calN)$ and $\Log{(\bcalY, \bcalU)}{\G_\Y,\G_\U}$ is the logarithmic map that maps the attractors $\G_\Y\in \calM$ and $\G_\U\in\calN$ from the manifold composite $\calM \times \calN$ to the tangent space $\calT_{(\Y_l,\U_l)}(\calM\times \calN)$ at each time-step. 

As an illustrative example, consider the pose of the end-effector of a robot, which can be represented as the Cartesian product of the hypersphere $\manS{3}$ and 3D-Euclidean space $\manR{3}$, \ie $\mathcal{H} = \manS{3} \times \manR{3}$. It is worth mentioning that the pose of the end-effector of a robot can be alternatively represented as a homogeneous transformation matrix $\bm{H} \in \mathcal{SE}(3)$ using the Lie group theory formulation~\cite{sola2018micro}; however, in this work, we exploit the Cartesian product property of Riemannian
manifolds.

\begin{remark}
\label{rm:stab_composite}
The stability of manifold composites \ac{gadmp} formulation in \eqref{eq:DMP_Comp_accel} and \eqref{eq:DMP_Comp_velocity} can be straightforwardly proven by applying Theorem~\ref{th:stab_rdmp} separately to $\calM$ and $\calN$.
\end{remark}

\begin{figure*}[t]
	\centering
	\def\svgwidth{\linewidth}
	{\fontsize{8}{8}
		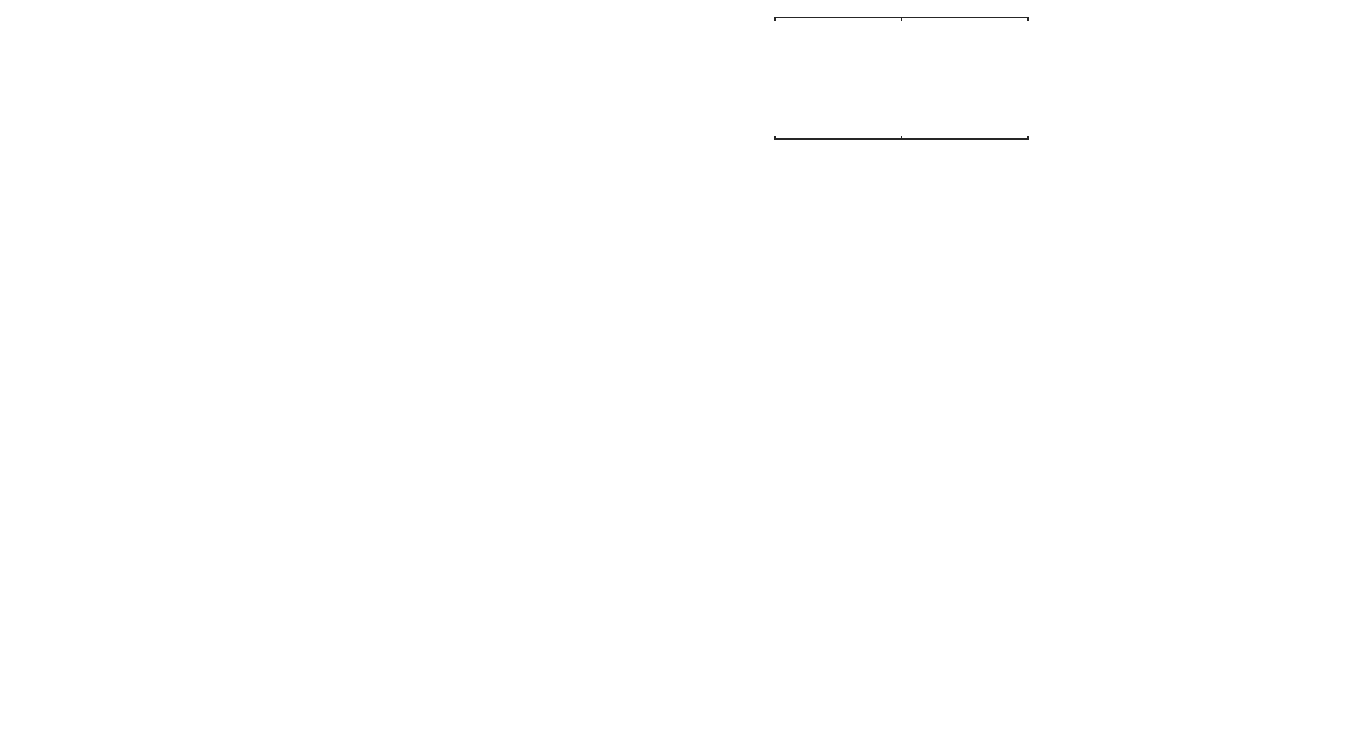}
	\caption{Illustrates the performance of \ac{gadmp} when executing Riemannian LASA dataset. \ordinal{1}{st} \emph{row}: Euclidean 2D trajectory. \ordinal{2}{nd} \emph{row}: Unit quaternion trajectory. \emph{\ordinal{3}{rd} row:} Rotation matrix trajectory. \ordinal{4}{th} \emph{row}: \ac{spd} trajectory. \ordinal{1}{st} \emph{column}: Trajectories from different manifolds. \ordinal{2}{nd} \emph{column}: first-derivative in different manifolds. \emph{\ordinal{3}{rd} column:} The distance in each manifold between the demonstration and the \ac{gadmp} reproduction. \ordinal{4}{th} \emph{column}: The Cartesian representation of the \ac{gadmp} reproduction. In \ordinal{1}{st} and \ordinal{2}{nd} \emph{columns}, dashed lines represent demonstration data while colored solid lines represent the \ac{gadmp} results. }
	\label{fig:lasa_experiment}
\end{figure*}

\section{Validation}
\label{sec:exper}
We validated the proposed \ac{gadmp} in simulation as well as in real setups. More in \change{detail}, we performed the following evaluations:
\begin{itemize}
    \item In simulation:
    \begin{itemize}
        \item[--] We augmented two public datasets; 2D-LASA handwriting dataset \cite{Khansari2011learning} and 2D-Letters handwriting dataset \cite{Calinon2020Gaussians} with data samples from three Riemannian manifolds (unit quaternion, rotation matrix, and symmetric and positive definite matrix).
    	\item[--] We compared  \ac{gadmp} with the baseline approaches~\cite{koutras2020correct} and~\cite{abudakka2020Geometry}.
    	\item[--] Learning manipulability ellipsoids and position by learning $\manR{2}\times \spd{2}$ with \ac{gadmp}.
    	\item[--] Goal switching simulation.
    \end{itemize}
    \item In real experiment:
    \begin{itemize}
        \item[--] Refilling a watering can by learning $\manR{3}\times \manS{3}\times \spd{3}$ with \ac{gadmp}.
        \item[--] Picking from different boxes task by learning $\manR{3}\times \spd{3}$ with \ac{gadmp}.
    \end{itemize}
\end{itemize}

We have created one by modifying the 2D-LASA and the 2D-Letters datasets. Mainly, we extended both datasets to include $\manS{3},\manSO{3}$, and $\spd{2}$ along with the original $\manR{2}$. 
The 2D-LASA handwriting dataset contains $30$ classes of $2$D Euclidean motions starting from different initial points and converging to the same goal $[0,0]\trsp$. Each motion is demonstrated $7$ times. A demonstration has exactly $1000$ samples and includes position, velocity, and acceleration profiles. 
On the other hand, the 2D-Letters handwriting dataset contains $26$ letters of $2$D Euclidean motions starting from different initial points and ending to different goals. Each motion is demonstrated $10$ times. A demonstration has exactly $200$ samples and includes position, velocity, and acceleration profiles.

The key idea to generate Riemannian data \change{from} Euclidean points is to consider each demonstration as an observation of a motion in the tangent space of a given Riemannian manifold. This allows us to use the exponential map to project the motion onto the manifold. In both datasets, demonstrations are in 2D ($xy$-plane), however, in order to create the 3D tangent space for both $\manS{3}$ and $\manSO{3}$, we added a $z$-axis to each demonstration as an average of $x$- and $y$-axes. As a result, we obtain $\manS{3}$ and $\manSO{3}$ demonstrations for each demonstration from both datasets. 

In order to create \ac{spd} training data profiles, we followed different strategies and used the 2D-LASA dataset to generate covariance matrix profiles and the 2D-Letters dataset to generate manipulability profiles. More in \change{detail}, we first fit a \ac{gmm} for each class of the 2D-LASA dataset. We then used \ac{gmr} to retrieve a $2\times 2$ covariance matrix profile. This covariance matrix profile served as SPD training data for \ac{gadmp}. Instead, for the 2D-Letters dataset, we placed the base of a 3-\ac{dof} 2D-manipulator at $[0,0]\trsp$, and determined the manipulability profile of the manipulator while it tracks the Cartesian trajectory of each demonstration. This manipulability profile served as SPD training data for \ac{gadmp}.

\subsection{Validation using Riemannian LASA dataset}
In order to validate the accuracy of the proposed unified \ac{dmp} formulation, we created $4$ tests in $4$ different manifolds, $\bcalP \in \manR{2}$, $\bcalQ \in \manS{3}$, $\bcalR \in \manSO{3}$, and $\bcalC \in \spd{2}$. These are illustrated in \figref{fig:lasa_experiment} where each row corresponds to a particular manifold. The leftmost column of the figure represents the evolution of the elements of the profile over time\footnote{As \ac{spd} matrices are symmetric, and for visualization purposes, in this figure we visualize the \ac{spd} by plotting the corresponding Mandel representation.}. Dashed black lines represent the demonstration and colored lines the reproduction of \ac{gadmp}. The second column corresponds to the \ordinal{1}{st}-time-derivative of the profiles in each manifold, while the \ordinal{3}{rd} column shows the error or the distance between the \ac{gadmp} profile and the demonstration profile for each manifold. The last column (rightmost) shows \change{what} the profile looks like in Cartesian space. In \change{the} case of $\manS{3}$, we rotate the 3D-frame of the 3D-Cartesian profile of the G-shape, while in $\manSO{3}$ we show the frame rotating around $[0,0,0]\trsp$. In \change{the} case of the $\spd{2}$, we illustrated the covariance matrices over the 2D-Cartesian profile of the G-shape. The results shown in this figure demonstrate the accuracy of the proposed \ac{gadmp} to reproduce the desired trajectory profiles in different manifolds.

\subsection{Comparison with \texorpdfstring{\cite{koutras2020correct}}{}}\label{subsec:greek_comparison}

\begin{figure}[t]
	\centering
	\def\svgwidth{\linewidth}
	{\fontsize{8}{8}
		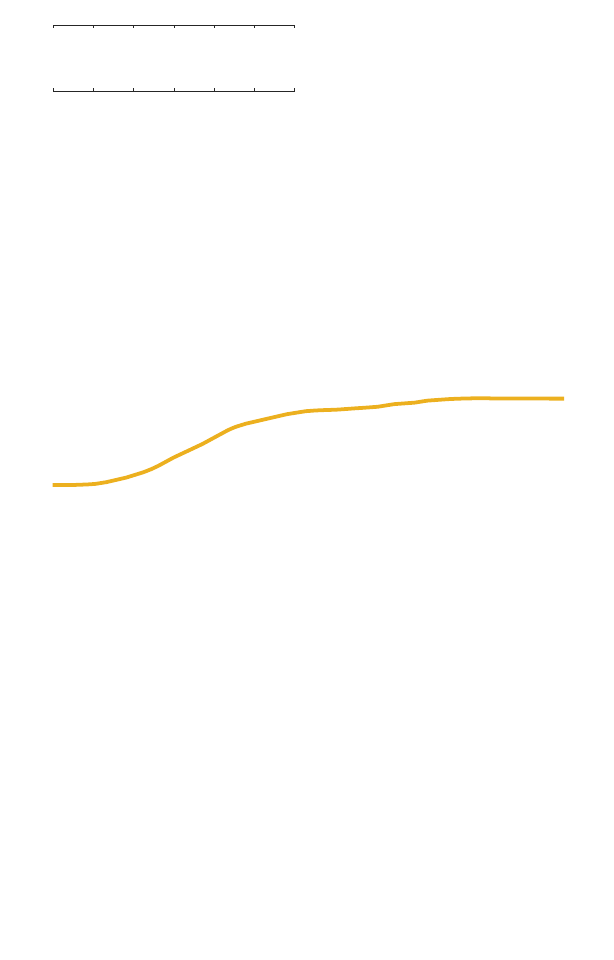}
	\caption{\ac{gadmp} execution of the same unit quaternion trajectory tested in \cite{koutras2020correct}. The first three \emph{rows} show the error between the current unit quaternion and the goal (\emph{left}) and new goal (\emph{right}). The \emph{bottom} four \emph{rows} show the evolution of each unit quaternion element, over time, toward \change{the} goal and new goal. Dashed black lines represent information related to \change{the} demonstration trajectory.}
	\label{fig:corlPaper}
\end{figure}

\begin{figure}[t]
	\centering
	\def\svgwidth{\linewidth}
	{\fontsize{8}{8}
		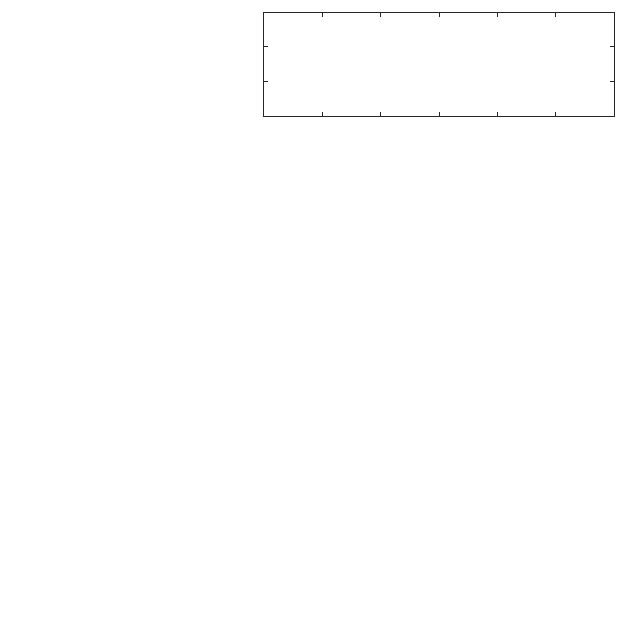}
	\caption{Comparison between the proposed \ac{gadmp} and \cite{koutras2020correct}. The first three \emph{rows} show more stable starting using \ac{gadmp}. \emph{Bottom:} Compares the mean error of \ac{gadmp} (in red) and \cite{koutras2020correct} (dashed black lines).}
	\label{fig:corlPaperERRcompare}
\end{figure}

The proposed \ac{gadmp} is rigorously derived in Sec.~\ref{sec:dmpFormulation} starting from a generic second-order dynamics evolving on a manifold. Therefore, our formulation is mathematically correct and it does not exhibit the oscillatory behaviors described in ~\cite{koutras2020correct}. In addition to the mathematical derivation, we provide in this simulation an experimental comparison to support our claim.

More in detail, we compared our \ac{gadmp} against the quaternion-based \ac{dmp} proposed in\footnote{We thank Leonidas Koutras for sharing with us the implementation and test trajectory of their work in \cite{koutras2020correct}.}~\cite{koutras2020correct}. We used the same simulated unit quaternion trajectory, where the initial and final quaternions are $\bm{Q}_0 = [-0.0092\ -0.7126\ 0.7015\ 0.0090]\trsp$ and $\bm{Q}_g = [0.8104\ 0.3364\ 0.2141\ 0.4293]\trsp$. Moreover, we used the same \ac{dmp} parameters, \eg $\alpha_z = 60$, $N = 60$, and $\alpha_x = 4.6052$. Top-left column of \figref{fig:corlPaper} shows the evolution of the quaternion error computed between the current (from \ac{gadmp}) and goal quaternions through $e_\bcalQ = 2\Log{\bcalQ}{\bm{Q}_g}$. The top-right column shows the evolution of the error toward a new goal $\bm{Q}_g^{new} = [0.7442\ 0.5414\ -0.0343\ 0.3897]\trsp$. The bottom 4 plots, show the evolution of the trajectories of unit quaternion elements toward the original goal and the new one. This figure shows the accuracy of the proposed \ac{gadmp} to encode and execute a challenging unit quaternion trajectory. Moreover, it is clear that \ac{gadmp} successfully performs a goal-switching task.

\Figref{fig:corlPaperERRcompare} compares the accuracy of our \ac{gadmp} with the approach proposed in \cite{koutras2020correct}. The \emph{bottom} plot shows that the proposed \ac{gadmp} is more accurate. 

Furthermore, the computational complexity during execution, particularly in terms of step time, remains compatible with control frequencies. Specifically, the means of the computational cost exhibited by \cite{koutras2020correct} and \ac{gadmp} at each control cycle are $0.04\,$ms and $0.1\,$ms, respectively. We also consider a baseline approach that uses the classical DMP and performs an extra normalization of the output. For the baseline, the mean computational cost for integrating and normalizing the output to reproduce a unit quaternion is $0.008\,$ms per time step. This indicates that all considered approaches can comfortably operate at frequencies exceeding $1\,$kHz, ensuring real-time responsiveness in robotic control applications.

\subsection{Comparison with \texorpdfstring{\cite{abudakka2020Geometry}}{}}
\label{subsec:icra_comparison}
To illustrate the difference between our new formulation in \eqref{eq:DMP_RM_accel}--\eqref{eq:DMP_RM_velocity} and our previous formulation described in \cite{abudakka2020Geometry}, where parallel transport was employed, we have conducted an experiment where both approaches executed 20 $\spd{2}$ trajectories of the modified Riemannian LASA dataset (\secref{sec:exper}).
\begin{figure}[t]
	\centering
	\def\svgwidth{\linewidth}
	{\fontsize{8}{8}
		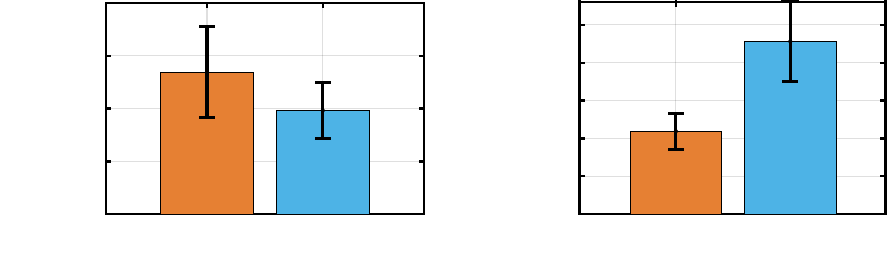}
	\caption{Comparison between the proposed \ac{gadmp} and our previous approach using parallel transport \cite{abudakka2020Geometry}. Both approaches executed 20 $\spd{2}$ trajectories of the modified Riemannian LASA dataset. \emph{Left}: The error distance between the demonstration and the reproduction. \emph{Right}: The computational cost in milliseconds per control cycle.}
	\label{fig:compICRA}
\end{figure}
Figure~\ref{fig:compICRA} shows bar plots for computational time required for both approaches to learn and execute complete trajectories, and the log-Euclidean distance \cite{Jayasumana2015} between the generated \ac{spd} profiles and the ground truth demonstrations. 
	
\begin{figure}[t]
	\centering
	\def\svgwidth{\linewidth}
	{\fontsize{8}{8}
		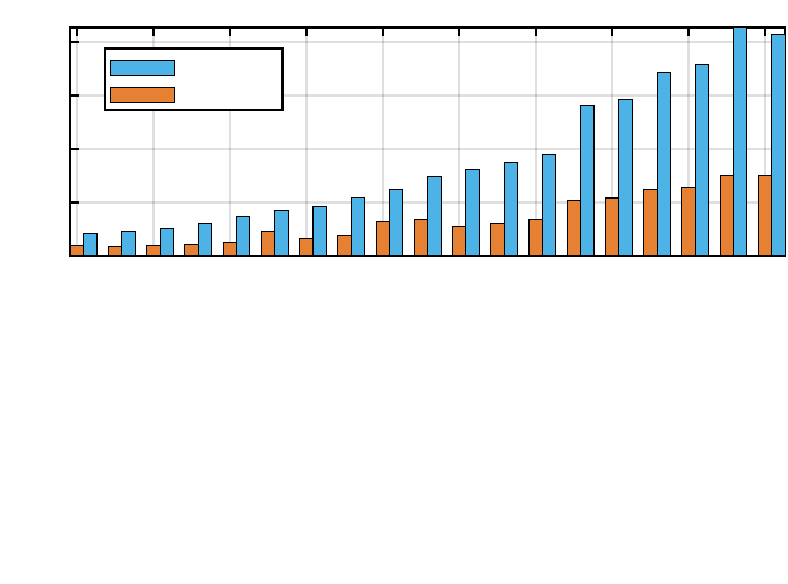}
	\caption{Comparison between the proposed \ac{gadmp} and our previous approach using parallel transport \cite{abudakka2020Geometry}. Both approaches executed 19 $\spd{m}$ trajectories, where $m=2,\dotsc, 20$. \emph{Top}: The computational cost in milliseconds per control cycle. \emph{Bottom}: The error distance between the demonstration and the reproduction. }
	\label{fig:compICRA_HD}
\end{figure}

Results in Fig.~\ref{fig:compICRA} show that employing parallel transport provides slightly more accurate results, as evidenced by the reduced log-Euclidean distance from the ground truth demonstrations. However, this improvement comes at a significant computational cost, as indicated by the increased computational time required for this approach. For instance, the mean of the computational cost exhibited by \cite{abudakka2020Geometry} and \ac{gadmp} at each control cycle are $0.09\,$ms and $0.04\,$ms, respectively. 

In Fig.~\ref{fig:compICRA_HD} we observe how this computational cost increases exponentially with the approach in \cite{abudakka2020Geometry} as problem dimensions increase. Though \cite{abudakka2020Geometry} exhibits a slight improvement in accuracy, this must be weighed against its heightened computational demands. In this example, we executed both approaches, in \cite{abudakka2020Geometry} and \ac{gadmp}, over 19 \ac{spd} trajectories with dimensions ranging from $\spd{2}$ to $\spd{20}$, providing a comprehensive comparison.

This trade-off between accuracy and computational efficiency is an important consideration in the selection of the appropriate formulation for specific applications. For tasks where computational resources are abundant and accuracy is paramount, the parallel transport approach may be preferred. However, the new formulation offers a more efficient alternative without penalizing the accuracy for real-time applications or scenarios with limited computational resources.
Finally, it is important to note that, while the approach in \cite{abudakka2020Geometry} is specifically designed for \ac{spd} matrices, our \ac{gadmp} framework is applicable to any Riemannian manifold.

\subsection{Learning manipulability ellipsoids}
\label{sec:learManElli}
The manipulability of a robotic arm provides an analytical way to evaluate the manipulator's ability to change its end-effector pose from a certain joint configuration. Manipulability can be illustrated as an ellipsoid in 2- or 3-D Euclidean space.
Mathematically, the manipulability of a robotic arm is computed from the forward kinematics 
\begin{equation}
	\bcalPd = \bm{\mathbb{J}} \bcalJd,
	\label{eq:invVel}
\end{equation}
that relates task velocity $\bcalPd \in \manR{m}$  and the joint velocity $\bcalJd \in \manR{n}$ through the Jacobian matrix $\bm{\mathbb{J}}\in \manR{m\times n}$. 
By considering, in~\eqref{eq:invVel}, only the joint velocity with unit norm, \ie $\norm{\bcalJd}=\bcalJd\trsp \bcalJd=1$, we obtain
\begin{equation}
	\bcalJd\trsp \bcalJd = \bcalPd\trsp(\bm{\mathbb{J}}^\dag)\trsp \bm{\mathbb{J}}^+\bcalPd = \bcalP\trsp\left(\bm{\mathbb{J}} \bm{\mathbb{J}}\trsp\right)^{\dag} \bcalPd,
\end{equation}
which defines a point on the surface of an ellipsoid in the end-effector velocity space. 
The \ac{spd} matrix $\bm{\Upsilon}=\left(\bm{\mathbb{J}} \bm{\mathbb{J}}\trsp\right)^{\dag}\in \spd{m} $, called manipulability ellipsoid, gives an intuition of the directions where the manipulator can move its end-effector at large/small velocities.

\begin{figure}[t]
	\centering
	\def\svgwidth{\linewidth}
	{\fontsize{8}{8}
		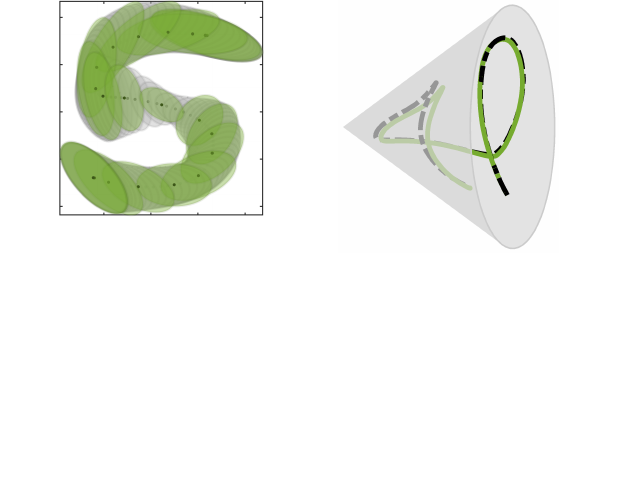}
	\caption{\emph{Top-Left}: The Cartesian trajectory (in centimeters) executed by the 5-\ac{dof} manipulator (black dots), the demonstrated manipulability profile (gray ellipses), and the manipulability profile learned by \ac{gadmp} (green ellipses), shown at different times during the execution of the task. \emph{Top-Right}: Representation of \ac{spd} manifold (gray cone) containing the demonstrated (dashed black line) and learned (green solid line) manipulability profiles. \emph{Bottom}: Variation of demonstrated (gray ellipses) and learned (green ellipses) manipulability profiles over time.}
	\label{fig:manEllip}
\end{figure}

Here we propose to use a toy example similar to the one in \cite{Rozo2017} to evaluate our \ac{gadmp} formulation while operating \ac{spd} data profiles.
One demonstration $\bm{\Xi} = \{t_{l}, \bm{\Upsilon}_{l}\}_{l=1}^T$ is obtained by performing a tracking task with a 3-\ac{dof} manipulator. Let us call $\bcalP$ the Cartesian position trajectory of the robot end-effector. The desired position trajectory $\bcalPhat$ is then tracked by a 5-\ac{dof} robot. The force $\bcalF$ needed to perform the tracking task is computed using the following control law originally proposed in~\cite{Rozo2017} 
\begin{equation}
	\bm{\tau}_d = \bm{\mathbb{J}}\trsp\bcalF-\left(\bm{I}-\bm{\mathbb{J}}\trsp\bm{\bar{\mathbb{J}}}\trsp\right)\alpha \triangledown g_t(\bcalJ); \quad \alpha>0,
	\label{eq:manip_tracking}
\end{equation}
where $\bm{\bar{\mathbb{J}}}$ is the inertia-weighted pseudo-inverse of $\bm{\mathbb{J}}$ and $\bm{\tau}_d$ is the desired joint torque. The cost function $g_t(\bcalJ)$ is defined as 
\begin{equation}
	\begin{aligned}
		g_t(\bcalJ)&=\text{log}\left(\text{det}\left(\frac{\bm{\hat{\Upsilon}}_t+\bm{\Upsilon}_{a,t}(\bcalJ)}{2}\right)\right)\\
		&\quad-\frac{1}{2}\text{log}\left(\text{det}\left(\bm{\hat{\Upsilon}}_t\bm{\Upsilon}_{a,t}(\bcalJ)\right)\right)
	\end{aligned},
\end{equation}
where $\bm{\Upsilon}_{a,t}(\bcalJ)$ are the actual and $\bm{\hat{\Upsilon}}_t$ the desired manipulability ellipsoids, respectively. $\bm{\hat{\Upsilon}}_t$ are generated using the proposed \ac{gadmp}. 

The results of this procedure, applied to track a 2-D S-shape Cartesian trajectory, \change{are} shown in Fig.~\ref{fig:manEllip}. 
Figure~\ref{fig:manEllip}(\emph{top-left}) shows that the desired manipulability profile (green ellipses) smoothly and accurately follows the demonstrated manipulability profile (gray ellipses) while the 5-\ac{dof} robot was performing the tracking task. Similar results are shown in Fig.~\ref{fig:manEllip}(\emph{bottom}), but considering the time evolution of desired and demonstrated manipulability ellipsoids. 
Figure~\ref{fig:manEllip}(\emph{top-right}) depicts the \ac{spd} manifold (a cone) and the geodesic curve of the desired and demonstrated manipulability profiles. The \ac{gadmp} successfully and accurately followed the demonstrated Cartesian trajectory along with the manipulability profile, in its composite Riemannian form $\manR{2}\times\spd{2}$, and converged to the goal.

\subsection{Goal switching}
\label{sec:goalSwitchingEXP}

\begin{figure}[ht]
	\centering
	\def\svgwidth{\linewidth}
	{\fontsize{8}{8}
		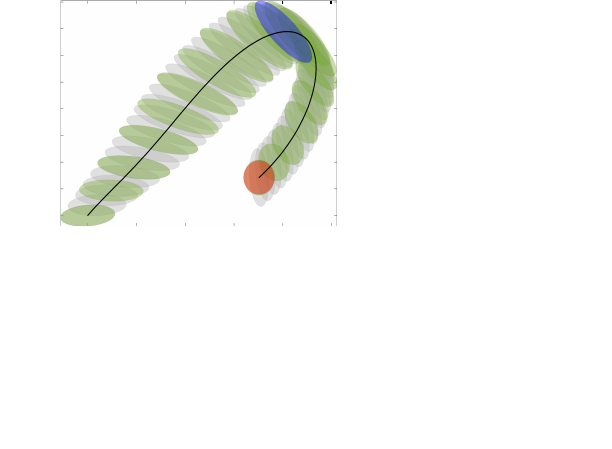}
	\caption{\ac{gadmp} adapts the stiffness profile to a new goal using the mechanism of goal switching \eqref{eq:riem_goal_switch}. Gray ellipsoids represent the demonstrated stiffness profile, green ones are the result of \ac{gadmp}, the blue one indicates the instant where goal switching occurred, and the red one denotes the new goal ellipsoid.  \emph{Top-Left}:  The evolution of \ac{gadmp} over a Cartesian trajectory. \emph{Bottom}: The evolution of \ac{gadmp} over time. \emph{Top-Right}: The evolution of the spring forces while tracking the Cartesian trajectory.}
	\label{fig:stifElliGoal}
\end{figure}

\begin{figure}[th]
	\centering
	\def\svgwidth{\linewidth}
	{\fontsize{8}{8}
		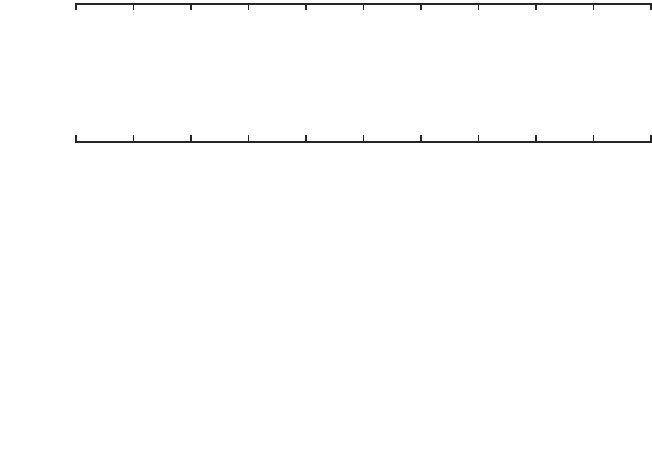}
	\caption{\emph{Top}: The Log-Euclidean distance between \ac{gadmp} evolution and the goal in both cases; reproduction (dashed black lines), adaptation using goal switching (red solid line). \emph{Bottom}: The element of stiffness profile in reproduction (dashed black lines) and adaptation using goal switching (colored solid lines).}
	\label{fig:stifDistGoal}
\end{figure}

In order to evaluate the proposed \ac{gadmp} formulation characteristics under goal switching, we used it to drive an \ac{msd}, with \change{a} designed variable stiffness profile, along a specific Cartesian trajectory. The variable stiffness profile is designed, such that, it starts with, horizontally-aligned stiffness ellipsoid, $[622.9934\ \   39.9577;\ 39.9577\ \   79.5444]$, then we rotated it gradually $90^\circ$, through ${\mathbf{R}^\trsp\bcalK\mathbf{R}}$ ($\mathbf{R}$ is a rotation matrix), until it ends up with, vertically-aligned stiffness ellipsoid, $[79.5444\ \   -39.9577;\ -39.9577\ \   588.2443]$. This stiffness profile $\bcalK\in\spd{2}$ is our demonstration, the gray ellipsoids in Fig.~\ref{fig:stifElliGoal}(\emph{top-left}), along with the Cartesian trajectory $\bcalP\in\manR{2}$, solid black curve. In this simulation, \ac{gadmp} encodes the composite Riemannian manifolds $\manR{2}\times\spd{2}$.

During the execution, we estimated the spring forces $\bm{f}^s$ while tracking the Cartesian trajectory. The \ac{gadmp} reproduction, in the first execution, has been successfully converged to the original goal, dashed lines in Fig.~\ref{fig:stifDistGoal}(\emph{bottom}). In the second execution, we switched to a new stiffness goal $[200\ \ 0;\ 0\ \ 200]$, red ellipsoid in Fig.~\ref{fig:stifElliGoal}, at the middle of the execution. From Fig.~\ref{fig:stifDistGoal}(\emph{top}), we can see the error between \ac{gadmp} stiffness result, at each time step, and the new stiffness goal converges to zero (the solid red line), which indicates that the \ac{gadmp} converges accurately to the new stiffness goal. 
 
\subsection{Robot experiments}\label{subsec:panda_experiments}
We evaluated the proposed approach on a $7$ \ac{dof} Franka Emika Panda robot with two experiments, namely picking from different boxes and refilling a watering can. In order to perform these tasks, the robot had to continuously modulate its position, orientation, stiffness, and/or manipulability. In real settings, orientation trajectories are often collected from demonstrations with a real robot. This requires a preprocessing step to extract unit quaternions \change{from} a trajectory of rotation matrices. The step is needed because the robot's forward kinematics is typically expressed as a homogeneous transformation matrix~\cite{siciliano2009robotics}. Numerical approaches to continuously compute quaternions from rotation matrices may return a quaternion at time $t$ and its antipodal at $t+1$, since antipodal quaternions represent the same rotation. The resulting discontinuity can be avoided by checking that the dot product $\boldfrak{q}_t\cdot \boldfrak{q}_{t+1} > 0$ and replacing $\boldfrak{q}_{t+1}$ with $-\boldfrak{q}_{t+1}$ otherwise.

\subsubsection{Refilling a watering can}
\begin{figure}[t]
	\centering
	\def\svgwidth{\linewidth}
	{\fontsize{8}{8}
		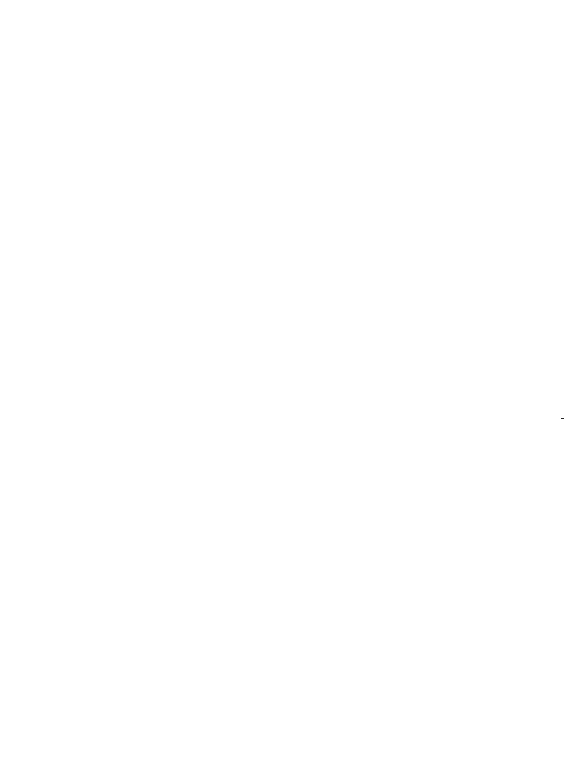}
	\caption{Results for the refill of a watering can experiment. \emph{Top}: The robot correctly performs the task. \emph{Bottom}: Position, orientation, and stiffness profiles.}
	\label{fig:water_experiment}
\end{figure}

In this experiment, the robot had to refill a watering can by immersing it in a tray full of water (see \figref{fig:water_experiment}). To perform the task, 
the robot was controlled using the Cartesian impedance control law
\begin{equation}
\begin{split}
	\bcalF_{p}&=\bcalK_{p} \left(\bcalP^{dmp}-\bcalP\right) + \bcalD_{p} \left(\bcalPd^{dmp}-\bcalPd\right), \\
	\bcalF_{o}&=\bcalK_{o}\, \Log{\bcalQ}{\bcalQ^{dmp}} + \bcalD_{o} \left(\bcalW^{dmp}-\bcalW\right),
	\end{split}
	\label{eq:robot_control}
\end{equation}
where the subscript $p$ indicates position and $o$ orientation. The measured end-effector position and orientation (unit quaternion) are indicated by $\bcalP$ and $\bcalQ$ respectively, and the \change{corresponding} linear and angular velocities are $\bcalPd$ and $\bcalW$.
The desired trajectories $\bcalP^{dmp}$ and $\bcalQ^{dmp}$, as well as the variable stiffness matrix $\bcalK_{p}$ and the desired velocities $\left(\bcalPd^{dmp} \text{ and } \bcalW^{dmp}\right)$, were generated with the proposed \ac{gadmp}. The orientation stiffness was kept constant at $\bcalK_{o}=150\,\bm{I}\,$Nm/rad.
The damping matrices $\bcalD_{p}$ and $\bcalD_{o}$ were computed from the respective stiffness matrices using the double diagonalization approach~\cite{albu2003cartesian}. The robot was controlled at $1\,$KHz using the joint torques
\begin{equation}
\boldfrak{\tau}_d = \bm{\mathbb{J}}\trsp	
\begin{bmatrix}
\bcalF_{p} \\ \bcalF_{o} 
\end{bmatrix},
\label{eq:jntImpd}
\end{equation}
where $\bm{\mathbb{J}}\trsp$ is the transpose of the manipulator Jacobian and the Cartesian forces $\bcalF_{p}$ and $\bcalF_{o}$ are defined as in~\eqref{eq:robot_control}.

Desired position, velocity, and stiffness profiles were learned using the proposed \ac{gadmp}. In order to estimate a variable stiffness profile, we collected $5$ kinesthetic demonstrations containing end-effector positions, velocities, accelerations, and sensed forces. These data were used through the interaction model proposed in \cite{AbuDakka2018} to estimate the variable stiffness profile shown in \figref{fig:water_experiment} (bottom). Positions and unit quaternion trajectories were learned from a single demonstration, obtained by averaging the $5$ used to obtain the stiffness profile. 

The results in \figref{fig:water_experiment} show that the proposed \ac{gadmp} formulation is capable of learning complex trajectories evolving on composite Riemannian manifolds $\manR{3}\times \manS{3}\times \spd{3}$ while fulfilling the underlying geometric constraints, \ie unit norm in variable orientation and symmetry and positive definiteness in variable stiffness profiles.

\begin{figure*}[t]
	\centering
	\def\svgwidth{\linewidth}
	{\fontsize{8}{8}
		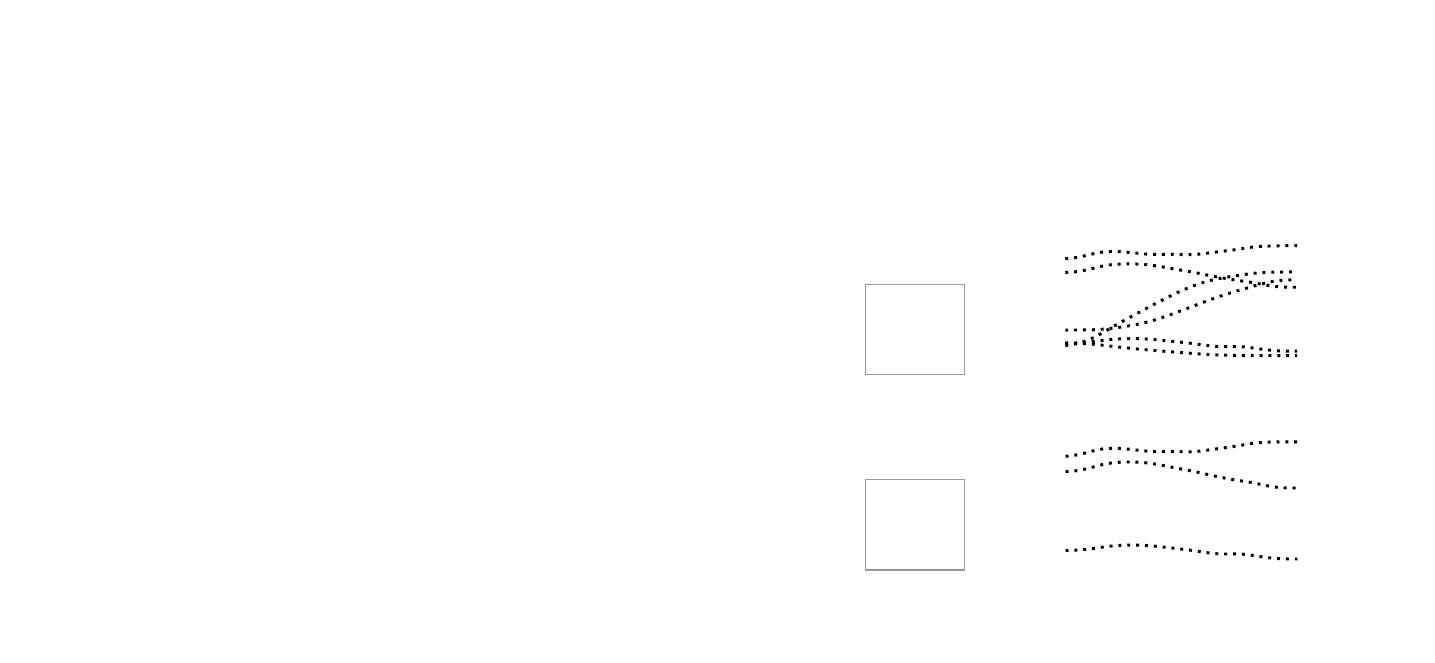}
	\caption{Results for the pick from different boxes experiment. \emph{Top}: Picking from the demonstrated box $1$. \emph{Middle}: Goal switching is used to pick from a new box $2$. \emph{Bottom}: Goal switching is used to pick from a new box $3$. In the $3$ cases, manipulability is controlled in the null-space of the position task to maintain a certain joint configuration during the motion. }
	\label{fig:pick_experiment}
\end{figure*}

\subsubsection{Pick from different boxes} 
In this experiment, the robot had to enter $3$ boxes placed at different locations, mimicking a pick from each of the boxes (see \figref{fig:pick_experiment}). The experiment was designed to show that geometry-aware \acp{dmp} can \textit{i)} effectively encode manipulability profiles and \textit{ii)} change the goal after the learning. 

We provided a kinesthetic demonstration to make the robot enter box $1$ while collecting end-effector position and joint trajectories. As detailed in \secref{sec:learManElli}, collected trajectories were used to learn position and manipulability profiles using geometry-aware \acp{dmp}. At run time, the robot was controlled using the control law~\eqref{eq:manip_tracking} to track the \ac{dmp} position as main task and to exploit its redundant \ac{dof} to follow the desired manipulability profile. As shown in \figref{fig:pick_experiment} (top), the robot followed accurately both position and manipulability profiles and successfully entered box $1$.

In order to experimentally verify the generalization capabilities of geometry-aware \acp{dmp}, we repeated the experiment by entering two boxes placed at different locations wrt box $1$. To measure the new goal, we manually placed the robot inside the boxes and stored its end-effector position. As shown in \figref{fig:pick_experiment} (middle)--(bottom), the robot reached the new position goals inside box $2$ and $3$. As already mentioned, the manipulability profile was tracked in the null-space of the position task, which introduces an error between the planned and executed manipulability profiles. However, in this task, null-space tracking was sufficient to preserve a joint configuration that let the robot enter boxes $2$ and $3$ without collision. 

Overall, the results in \figref{fig:pick_experiment} show that the proposed \ac{gadmp} formulation is capable of learning complex trajectories evolving on the composite Riemannian manifold $\manR{3}\times \spd{3}$ while fulfilling the underlying geometric constraints, \ie symmetry and positive definiteness in variable manipulability profiles.

\section{Conclusion}
\label{sec:concl}

In this paper, we have exploited Riemannian geometry to derive a new formulation of \ac{dmp} that is capable of learning and reproducing robot skills evolving on any Riemannian manifold. Our new formulation, \acf{gadmp}, is manifold independent and allows us to treat data belonging to different manifolds in a unified manner. It also preserves the underlying geometric constraints during both learning and reproduction without pre- or post-processing of the data. Moreover, it preserves the properties of the classical \ac{dmp} formulation such as convergence to a given target and the possibility to change the target at run-time (goal switching). 

\ac{gadmp} has been extensively validated through multiple simulation examples and two experiments on a real robotic manipulator. For simulation, we augmented two Euclidean datasets (2D-Letters and LASA handwriting) with data samples from three Riemannian manifolds ($\manS{3}$, $\manSO{3}$, and $\spd{2}$). We showed that \ac{gadmp} can accurately learn profiles evolving on such manifolds while converging to a (possibly changing) goal.
Moreover, a comparison with a baseline approach was conducted on a unit quaternion trajectory. In this case, \ac{gadmp} shows improvement by avoiding slight jumps at the beginning of the trajectories. Finally, real experiments show the effectiveness of \ac{gadmp} in encoding data from manifolds such as orientation, and \ac{spd} matrices.

In the future, we propose to integrate our approach with iterative learning algorithms---for example iterative learning control---in order to adapt to different situations and perform more complex tasks such as physical interaction control. Moreover, extending exploration-based learning methods to \change{Riemannian} manifolds is an open research problem. These methods are crucial when a robot needs to significantly adapt its behavior to a new situation by considering the data directly on its corresponding manifold. This will \change{allow} us to successfully exploit \acp{gadmp} in a large diversity of task situations.

\appendix
\section{Characterization of Used Manifolds}
%

\subsection{The \texorpdfstring{\ac{spd}}{} manifold \texorpdfstring{$\spd{m}$}{}}
\label{sec:manifoldSPD}

As early mentioned, \ac{spd} matrices is important in robotics as it encapsulate different types of data. The space $\spd{m}$ is defined  as the space of $m\times m$ \acl{spd} matrices. This space is not closed under scalar product and addition \cite{Pennec2006}, thus, we cannot use classical Euclidean arithmetic operators to manipulate these matrices. Alternatively, we can equip \ac{spd} matrices with A Riemannian metric in order to form a Riemannian manifold \cite{Pennec2006}. 

Note that the space $\spd{m}$ can be represented as the interior of a convex cone embedded in its tangent space of symmetric $m \times m$ matrices $\sym{m}$.

For $\bm{Q}, \bm{U}\in \spd{m}$ and $\bm{v}\in \calT_{\bm{U}}\spd{m}$, the logarithmic and exponential maps \eqref{eq:LOG} and \eqref{eq:EXP} can be defined as in  \cite{Pennec2006}
\begin{eqnarray}
	\bm{v} &=& \text{Log}_{\bm{U}}(\bm{Q}) = 	\bm{U}^\frac{1}{2}\text{logm}\Big(\bm{U}^{-\frac{1}{2}}\bm{Q}\bm{U}^{-\frac{1}{2}}\Big)\bm{U}^\frac{1}{2},
	\label{eq:logSPD} \\
    \bm{Q} &=& \text{Exp}_{\bm{U}}(\bm{v}) = \bm{U}^\frac{1}{2}\text{expm}\Big(\bm{U}^{-\frac{1}{2}}\bm{v}\bm{U}^{-\frac{1}{2}}\Big)\bm{U}^\frac{1}{2},\label{eq:expSPD}
	\end{eqnarray}
where $\text{logm}(\cdot)$ and $\text{expm}(\cdot)$ are the matrix logarithm and exponential functions.

\subsection{The unit $m$-sphere manifold \texorpdfstring{$\manS{m}$}{}}
\label{sec:manifoldSd}
$\manS{m}$ is a topological space embedded in $\manR{m+1}$ Cartesian space, where ${\manS{m}=\left\{\bm{X}\in\manR{m+1}:\norm{\bm{X}}=1\right\}}$. For $\bm{Q}, \bm{U}\in \manS{m}$ and $\bm{v},\bm{r}\in \calT_{\bm{U}}\manS{m}$ then, the logarithmic and exponential maps \eqref{eq:EXP} and \eqref{eq:LOG} are defined as in \cite{rentmeesters2011gradient}
\begin{eqnarray}
	\bm{v} &=& \Log{\bm{U}}{\bm{Q}} = \frac{\bm{Q}-(\bm{U}\trsp\bm{Q})\bm{U}}{\norm{\bm{Q}-(\bm{U}\trsp\bm{Q})\bm{U}}}d(\bm{U},\bm{Q}),	\label{eq:logSm}\\
	\bm{Q} &=& \Exp{\bm{U}}{\bm{v}} = \bm{U}\cos(\norm{\bm{v}})+\frac{\bm{v}}{\norm{\bm{v}}}\sin(\norm{\bm{v}}),\label{eq:expSm}
\end{eqnarray}
where $d(\bm{U},\bm{Q})\equiv\arccos(\bm{Q}\trsp\bm{U})$ defines the geodesic distance between $\bm{Q}$ and $\bm{U}$. 

\subsection{The unit quaternions group \texorpdfstring{$\manS{3}$}{}}
\label{sec:manifoldS3}

One way to describe the robot's end-effector orientation, in 3D-space, is to use unit quaternion representation. For $\bm{Q}, \bm{U}\in \manS{3}$ and $\bm{v},\bm{r}\in \calT_{\bm{U}}\manS{3}\equiv \manR{3}$, where $\manS{3}$ is a unit sphere in $\manR{4}$, $\bm{Q}=\nu_q+\bm{u}_q$, $\nu_q\in\manR{}$, and $\bm{u}_q\in\manR{3}$. 
The logarithmic and exponential maps \eqref{eq:LOG} and \eqref{eq:EXP} are 
\begin{eqnarray}
	\bm{v} &=& \begin{aligned}[t] &\Log{\bm{U}}{\bm{Q}} = \Log{}{\bm{Q}*\bm{\bar{U}}} \\&=  
    \begin{cases}
        \arccos(\nu) \frac{\bm{u}}{||\bm{u}||}, & \bm{u}\neq \bm{0} \\
        [0\;\; 0\;\; 0]{\trsp}, & \text{otherwise}.
    \end{cases}\end{aligned} \label{eq:logS3}\\
    \bm{Q} &=& \begin{aligned}[t] &\Exp{\bm{U}}{\bm{v}} \\&=
    \begin{cases}
        \left[\cos(||\bm{v}||)+\sin(||\bm{v}||)\frac{\bm{v}}{||\bm{v}||}\right] *\bm{U}, & \bm{v}\neq \bm{0} \\
        \left[1+[0\;\; 0 \,\;0]{\trsp}\right] *\bm{U}, & \text{otherwise}.
    \end{cases}  \end{aligned}
    \label{eq:expS3}
\end{eqnarray}
where $\bm{Q}*\bm{\bar{U}}=\nu+\bm{u}\in\manS{3}$, and $\bm{v}\in\manR{3}$ is treated as a quaternion with $\nu = 0$. 
\subsection{The special orthogonal group \texorpdfstring{$\manSO{m}$}{}}
\label{sec:manifoldSO}
$\manSO{m}$ is a subgroup of the orthogonal group $\mathcal{O}(m)$ where its determinant is $1$. Let us define $\R_1,\R_2 \in\manSO{m}$ and $\bm{v}\in\calT_{\R_1}\manSO{m}$, then the logarithmic and exponential maps \eqref{eq:EXP} and \eqref{eq:LOG} are defined as in \cite{rentmeesters2011gradient}
\begin{eqnarray}
	\bm{v} &=& \text{Log}_{\R_1}(\R_2) =\logm{\R_1\trsp\R_2},	\label{eq:logSOm} \\
	\R_2 &=& \text{Exp}_{\R_1}(\bm{v}) = \expm{\bm{v}} \R_1. \label{eq:expSOm}
\end{eqnarray}

\subsection{The rotation group \texorpdfstring{$\manSO{3}$}{}}
\label{sec:manifoldSO3}
Traditionally, orientations, in 3D-space, were represented through rotation matrices in $\manSO{3} = \{\R\in \manR{3\times 3} : |\R| = 1, \R\trsp\R = \R\R\trsp=\bm{I}\}$ which are widely used in robotics. Let us define $\R_1,\R_2 \in\manSO{3}$ and $\bm{v}\in\calT_{\R_1}\manSO{3}$, then \eqref{eq:LOG} will be \cite{murray2017mathematical}
\begin{eqnarray}
	\begin{aligned}\bm{v} &= \Log{\R_1}{\R_2} = \Log{}{\R_2\R_1\trsp} = \Log{}{\R} \\& = \begin{cases}
	    [0, 0, 0]\trsp, & \R = \bm{I} \\
		\omega = \theta \bm{n}, & \text{otherwise},
	\end{cases}
	\end{aligned}\label{eq:logSOm_}
\end{eqnarray}
where
\begin{equation*}
	\theta = \arccos{\left(\frac{\text{trace}(\R)-1}{2}\right)},\ 
	\bm{n}=\frac{1}{2\sin{(\theta)}}
	\begin{bmatrix}
		r_{32} - r_{23}\\
		r_{13} - r_{31}\\
		r_{21} - r_{12}
	\end{bmatrix}
\end{equation*}
and \eqref{eq:EXP} will be
\begin{eqnarray}
	\begin{aligned} \R_2 &=\Exp{\R_1}{\bm{[v]_\times}} \\& =  \left(\bm{I}+\sin(\theta)\frac{[\bm{v}]_\times}{||\bm{v}||} + (1-\cos(\theta))\frac{[\bm{v}]_\times^2}{||\bm{v}||^2}\right)\R_1,
	\end{aligned}\label{eq:expSOm_}
\end{eqnarray}

Note that the mappings in \eqref{eq:logS3}--\eqref{eq:expS3} and in \eqref{eq:logSOm_}--\eqref{eq:expSOm_} are computed using Lie group theory as unit quaternions and rotation matrices form a Lie group ~\cite{sola2018micro}. In particular, the mappings are based on the tangent space placed at the identity element (the so-called Lie algebra), and the product operations are used to parallel transport vectors from the Lie algebra to the tangent space placed at a different point ($\bm{U}$ or $\bm{R}_1$). We used the term Riemannian through the paper since every Lie group equipped with a Riemannian metric is a Riemannian manifold, but not vice versa.

\section*{Acknowledgements}
This work is supported in part by Basque Government (ELKARTEK) projects Proflow KK-2022/00024 and HELDU KK-2023/00055, in part by the European Union project INVERSE (GA No. 101136067), and in part by CHIST-ERA project IPALM (Academy of Finland decision 326304). Real experiments were conducted at the Department of Computer Science, University of Innsbruck, Austria.

\href{https://sites.google.com/view/abudakka/}{aaa}
%
%
%



\begin{thebibliography}{10}
	\expandafter\ifx\csname url\endcsname\relax
	\def\url#1{\texttt{#1}}\fi
	\expandafter\ifx\csname urlprefix\endcsname\relax\def\urlprefix{URL }\fi
	\expandafter\ifx\csname href\endcsname\relax
	\def\href#1#2{#2} \def\path#1{#1}\fi
	
	\bibitem{schaal1999Is}
	S.~Schaal, Is imitation learning the route to humanoid robots?, Trends in
	Cognitive Sciences 3~(6) (1999) 233--242.
	
	\bibitem{Billard2016Learning}
	A.~Billard, S.~Calinon, R.~Dillmann, Learning from Humans, 2016, pp.
	1995--2014.
	
	\bibitem{Ravichandar2020Recent}
	H.~Ravichandar, A.~S. Polydoros, S.~Chernova, A.~Billard, Recent advances in
	robot learning from demonstration, Annual Review of Control, Robotics, and
	Autonomous Systems 3~(1) (2020) 297--330.
	
	\bibitem{ijspeert2013}
	A.~J. Ijspeert, J.~Nakanishi, H.~Hoffmann, P.~Pastor, S.~Schaal, Dynamical
	movement primitives: learning attractor models for motor behaviors, Neural
	Computation 25~(2) (2013) 328--373.
	
	\bibitem{Khansari2011learning}
	S.~M. Khansari-Zadeh, A.~Billard, Learning stable non-linear dynamical systems
	with gaussian mixture models, IEEE Transactions on Robotics 27~(5) (2011)
	943--957.
	
	\bibitem{zadeh2014learning}
	S.~M. Khansari-Zadeh, A.~Billard, Learning control {L}yapunov function to
	ensure stability of dynamical system-based robot reaching motions, Robotics
	and Autonomous Systems 62~(6) (2014) 752--765.
	
	\bibitem{Ijspeert2002Learning}
	A.~J. Ijspeert, J.~Nakanishi, S.~Schaal, Learning rhythmic movements by
	demonstration using nonlinear oscillators, in: IEEE/RSJ International
	Conference on Intelligent Robots and Systems, Lausanne, Switzerland, 2002,
	pp. 958--963.
	
	\bibitem{Ude2014}
	A.~Ude, B.~Nemec, T.~Petri{\'c}, J.~Morimoto, Orientation in cartesian space
	dynamic movement primitives, in: IEEE International Conference on Robotics
	and Automation, Hong Kong, China, 2014, pp. 2997--3004.
	
	\bibitem{koutras2020correct}
	L.~Koutras, Z.~Doulgeri, A correct formulation for the orientation dynamic
	movement primitives for robot control in the cartesian space, in: Conference
	on Robot Learning, Osaka, Japan, 2020, pp. 293--302.
	
	\bibitem{huang2020toward}
	Y.~Huang, F.~J. Abu-Dakka, J.~Silv{\'e}rio, D.~G. Caldwell, Toward orientation
	learning and adaptation in cartesian space, IEEE Transactions on Robotics
	37~(1) (2020) 82--98.
	
	\bibitem{traversaro2016identification}
	S.~Traversaro, S.~Brossette, A.~Escande, F.~Nori, Identification of fully
	physical consistent inertial parameters using optimization on manifolds, in:
	IEEE/RSJ International Conference on Intelligent Robots and Systems, Daejeon,
	Korea, 2016, pp. 5446--5451.
	
	\bibitem{yoshikawa1985manipulability}
	T.~Yoshikawa, Manipulability of robotic mechanisms, The International Journal
	of Robotics Research 4~(2) (1985) 3--9.
	
	\bibitem{abudakka2021probabilistic}
	F.~J. Abu-Dakka, Y.~Huang, J.~Silv{\'e}rio, V.~Kyrki, A probabilistic framework
	for learning geometry-based robot manipulation skills, Robotics and
	Autonomous Systems 141 (2021) 103761.
	
	\bibitem{ikeura1995}
	R.~Ikeura, H.~Inooka, Variable impedance control of a robot for cooperation
	with a human, in: IEEE International Conference on Robotics and Automation,
	Nagoya, Japan, 1995, pp. 3097--3102.
	
	\bibitem{AbuDakka2018}
	F.~J. Abu-Dakka, L.~Rozo, D.~G. Caldwell, Force-based variable impedance
	learning for robotic manipulation, Robotics and Autonomous Systems 109 (2018)
	156--167.
	
	\bibitem{AbuDakka2015}
	F.~J. Abu-Dakka, B.~Nemec, J.~A. J{\o}rgensen, T.~R. Savarimuthu,
	N.~Kr{\"u}ger, A.~Ude, Adaptation of manipulation skills in physical contact
	with the environment to reference force profiles, Autonomous Robots 39~(2)
	(2015) 199--217.
	
	\bibitem{abudakka2020Geometry}
	F.~J. Abu-Dakka, V.~Kyrki, Geometry-aware dynamic movement primitives, in: IEEE
	International Conference on Robotics and Automation, Paris, France, 2020, pp.
	4421--4426.
	
	\bibitem{saveriano2019merging}
	M.~Saveriano, F.~Franzel, D.~Lee, Merging position and orientation motion
	primitives, in: IEEE International Conference on Robotics and Automation,
	Montreal, QC, Canada, 2019, pp. 7041--7047.
	
	\bibitem{saveriano2023dynamic}
	M.~Saveriano, F.~J. Abu-Dakka, A.~Kramberger, L.~Peternel, Dynamic movement
	primitives in robotics: A tutorial survey, The International Journal of
	Robotics Research 42~(13) (2023) 1133--1184.
	
	\bibitem{paraschos2013}
	A.~Paraschos, C.~Daniel, J.~R. Peters, G.~Neumann, Probabilistic movement
	primitives, in: Advances in Neural Information Processing Systems, Nevada,
	USA, 2013, pp. 2616--2624.
	
	\bibitem{calinon2014}
	S.~Calinon, D.~Bruno, D.~G. Caldwell, A task-parameterized probabilistic model
	with minimal intervention control, in: IEEE International Conference on
	Robotics and Automation, Hong Kong, China, 2014, pp. 3339--3344.
	
	\bibitem{pastor2009learning}
	P.~Pastor, H.~Hoffmann, T.~Asfour, S.~Schaal, Learning and generalization of
	motor skills by learning from demonstration, in: IEEE International
	Conference on Robotics and Automation, Kobe, Japan, 2009, pp. 763--768.
	
	\bibitem{abudakka2021Periodic}
	F.~J. Abu-Dakka, M.~Saveriano, L.~Peternel, Periodic dmp formulation for
	quaternion trajectories, in: IEEE International Conference of Advanced
	Robotics, Ljubljana, Slovenia, 2021, pp. 658--663.
	
	\bibitem{gousenbourger2019data}
	P.-Y. Gousenbourger, E.~Massart, P.-A. Absil, Data fitting on manifolds with
	composite b{\'e}zier-like curves and blended cubic splines, Journal of
	Mathematical Imaging and Vision 61~(5) (2019) 645--671.
	
	\bibitem{kim2017gaussian}
	S.~Kim, R.~Haschke, H.~Ritter, Gaussian mixture model for 3-dof orientations,
	Robotics and Autonomous Systems 87 (2017) 28--37.
	
	\bibitem{Zeestraten2017}
	M.~J. Zeestraten, I.~Havoutis, J.~Silv{\'e}rio, S.~Calinon, D.~G. Caldwell, An
	approach for imitation learning on riemannian manifolds, IEEE Robotics and
	Automation Letters 2~(3) (2017) 1240--1247.
	
	\bibitem{dodero2015kernel}
	L.~Dodero, H.~Q. Minh, M.~San~Biagio, V.~Murino, D.~Sona, Kernel-based
	classification for brain connectivity graphs on the riemannian manifold of
	positive definite matrices, in: IEEE International Symposium on Biomedical
	Imaging, Brooklyn, NY, USA, 2015, pp. 42--45.
	
	\bibitem{herath2017learning}
	S.~Herath, M.~Harandi, F.~Porikli, Learning an invariant hilbert space for
	domain adaptation, in: IEEE Conference on Computer Vision and Pattern
	Recognition, Honolulu, Hawaii, 2017, pp. 3845--3854.
	
	\bibitem{alexander2001spatial}
	D.~C. Alexander, C.~Pierpaoli, P.~J. Basser, J.~C. Gee, Spatial transformations
	of diffusion tensor magnetic resonance images, IEEE Transactions on Medical
	Imaging 20~(11) (2001) 1131--1139.
	
	\bibitem{Calinon2020Gaussians}
	S.~Calinon, Gaussians on {R}iemannian manifolds: Applications for robot
	learning and adaptive control, {IEEE} Robotics and Automation Magazine 27~(2)
	(2020) 33--45.
	
	\bibitem{Jaquier2017}
	N.~Jaquier, S.~Calinon, Gaussian mixture regression on symmetric positive
	definite matrices manifolds: Application to wrist motion estimation with
	semg, in: IEEE/RSJ International Conference on Intelligent Robots and
	Systems, Vancouver, Canada, 2017, pp. 59--64.
	
	\bibitem{pennec2006intrinsic}
	X.~Pennec, Intrinsic statistics on riemannian manifolds: Basic tools for
	geometric measurements, Journal of Mathematical Imaging and Vision 25~(1)
	(2006) 127--154.
	
	\bibitem{absil2009optimization}
	P.-A. Absil, R.~Mahony, R.~Sepulchre, Optimization algorithms on matrix
	manifolds, Princeton University Press, 2009.
	
	\bibitem{Pennec2006}
	X.~Pennec, P.~Fillard, N.~Ayache, A riemannian framework for tensor computing,
	International Journal of Computer Vision 66~(1) (2006) 41--66.
	
	\bibitem{frechet1948elements}
	M.~Fr{\'e}chet, Les {\'e}l{\'e}ments al{\'e}atoires de nature quelconque dans
	un espace distanci{\'e}, in: Annales de l'institut Henri Poincar{\'e},
	Vol.~10, 1948, pp. 215--310.
	
	\bibitem{fiori2022synthetic}
	S.~Fiori, I.~Cervigni, M.~Ippoliti, C.~Menotta, Synthetic nonlinear
	second-order oscillators on riemannian manifolds and their numerical
	simulation, Discrete and Continuous Dynamical Systems-B 27~(3) (2022)
	1227--1262.
	
	\bibitem{boumal2011discrete}
	N.~Boumal, P.-A. Absil, A discrete regression method on manifolds and its
	application to data on so(n), IFAC Proceedings Volumes 44~(1) (2011)
	2284--2289, 18th IFAC World Congress.
	
	\bibitem{Markus56}
	L.~Markus, Asymptotically autonomous differential systems, in: S.~Lefschetz
	(Ed.), Contributions to the Theory of Nonlinear Oscillations III, Princeton
	University Press, 1956, pp. 17--30.
	
	\bibitem{fiori2021manifold}
	S.~Fiori, Manifold calculus in system theory and control--fundamentals and
	first-order systems, Symmetry 13~(11) (2021).
	
	\bibitem{slotine1991applied}
	J.~Slotine, W.~Li, Applied nonlinear control, Prentice-Hall Englewood Cliffs,
	1991.
	
	\bibitem{pait2010properties}
	F.~{Pait}, D.~{Col{\'o}n}, Some properties of the riemannian distance function
	and the position vector x, with applications to the construction of lyapunov
	functions, in: IEEE Conference on Decision and Control, Atlanta, GA, USA,
	2010, pp. 6277--6280.
	
	\bibitem{sola2018micro}
	J.~Sola, J.~Deray, D.~Atchuthan, A micro lie theory for state estimation in
	robotics, arXiv preprint arXiv:1812.01537 (2018).
	
	\bibitem{Jayasumana2015}
	S.~Jayasumana, R.~Hartley, M.~Salzmann, H.~Li, M.~Harandi, Kernel methods on
	riemannian manifolds with gaussian rbf kernels, IEEE Transactions on Pattern
	Analysis and Machine Intelligence 37~(12) (2015) 2464--2477.
	
	\bibitem{Rozo2017}
	L.~Rozo, N.~Jaquier, S.~Calinon, D.~G. Caldwell, Learning manipulability
	ellipsoids for task compatibility in robot manipulation, in: IEEE/RSJ
	International Conference on Intelligent Robots and Systems, Vancouver,
	Canada, 2017, pp. 3183--3189.
	
	\bibitem{siciliano2009robotics}
	B.~Siciliano, L.~Sciavicco, L.~Villani, G.~Oriolo, Robotics: Modelling,
	Planning and Control, Springer, 2009.
	
	\bibitem{albu2003cartesian}
	A.~Albu-Schaffer, C.~Ott, U.~Frese, G.~Hirzinger, Cartesian impedance control
	of redundant robots: Recent results with the dlr-light-weight-arms, in: IEEE
	International Conference on Robotics and Automation, Taipei, Taiwan, 2003,
	pp. 3704--3709.
	
	\bibitem{rentmeesters2011gradient}
	Q.~Rentmeesters, A gradient method for geodesic data fitting on some symmetric
	riemannian manifolds, in: IEEE Conference on Decision and Control and
	European Control Conference, Orlando, FL, USA, 2011, pp. 7141--7146.
	
	\bibitem{murray2017mathematical}
	R.~M. Murray, Z.~Li, S.~S. Sastry, A mathematical introduction to robotic
	manipulation, CRC press, 2017.
	
\end{thebibliography}
\end{document}